\documentclass{article} 
\usepackage[final] {colm2026_conference} 

\usepackage{microtype}
\usepackage{hyperref}
\usepackage{url}
\usepackage{booktabs}
\usepackage{longtable}
\usepackage[inkscapelatex=false]{svg}

\usepackage[pass]{geometry}

\usepackage{lineno}
\usepackage{xcolor}
\usepackage{array}

\usepackage{listings}
\usepackage{tcolorbox}
\tcbuselibrary{listings}
\usepackage[T1]{fontenc}
\usepackage{float}
\usepackage{subcaption}

\definecolor{darkblue}{rgb}{0, 0, 0.5}
\hypersetup{colorlinks=true, citecolor=darkblue, linkcolor=darkblue, urlcolor=darkblue}

\title{\serum: State Extraction and Refinement for User Modeling}

\author{Andy J. Phu, Karin de Langis, James Mooney, Khanh Chi Le, Dongyeop Kang \\
Minnesota NLP Lab \\
University of Minnesota \\
Minneapolis, MN 55455, USA \\
\texttt{\{phu00003,moone174,dento019,le000422,dongyeop\}@umn.edu}
}

\usepackage{amsmath,amsfonts,bm}
\usepackage{xspace}

\def\eqref#1{equation~\ref{#1}}

\def\1{\bm{1}}

\DeclareMathAlphabet{\mathsfit}{\encodingdefault}{\sfdefault}{m}{sl}
\SetMathAlphabet{\mathsfit}{bold}{\encodingdefault}{\sfdefault}{bx}{n}

\newcommand{\serum}{\textsc{Serum}\xspace}

\begin{document}

\ifcolmsubmission
\linenumbers
\fi

\maketitle

\begin{abstract}
Agentic assistants capable of proactive, personalized interactions require structured models of user intent and workflow. 
However, building these models from raw, unstructured screen activity remains an open challenge. 
We present \serum, a multi-pass framework that extracts finite-state behavioral models directly from unstructured egocentric video using hierarchical VLM annotation. 
Processing screen recordings through a sliding window, \serum alternates between activity-recognition and intent-inference passes, with each pass refining labels using accumulated prior context to reduce hallucination and temporal conflation seen in single-pass annotation.
Synonymous states are then merged via sentence embeddings and human-calibrated thresholds into a compact, coherent taxonomy. 
We evaluate behavioral structure by fitting first-order Markov models over the resulting label sequences (both actions and intents) and measuring predictive accuracy against frequency baselines.  
Across 61 egocentric videos in four domains (coding, cooking, physical activities, and daily life), we find: (1) iterative label refinement converges to a stable state vocabulary, which we term \textit{schematic equilibrium}, after several passes; (2) normalized Markov models achieve substantially lower perplexity and higher action predictions than frequency baselines, with the largest gains on structured tasks like coding; and (3) human annotators rate final-pass labels as accurate and meaningfully improved over first-pass labels. 
To our knowledge, \serum is the first system to produce interpretable process models from unstructured egocentric screen video without manual annotation, opening a scalable pathway for user modeling and behavioral understanding in the wild. Our demo, code, and results are publicly available \footnote{\url{https://minnesotanlp.github.io/SERUM-web/}}

\end{abstract}


\section{Introduction}

Proactive AI assistants need structured models of user behavior — compact representations of how people move through goal-directed activity over time. Rich egocentric footage has been scarce before platforms like YouTube, and Twitch, .
converting raw video into structured behavioral models is non-trivial. Existing activity recognition methods either depend on fixed hand-crafted taxonomies~\citep{damen2022epickitchens100,grauman2022ego4d} or require expensive frame-level annotation. Process mining produces elegant behavioral models from event logs~\citep{vanderaalst2012manifesto,vanderaalst2016book}, but assumes activity labels already exist. Neither path applies to unstructured, open-ended video.

We ask: \emph{can we extract interpretable, structured models of user behavior directly from raw egocentric video, without a predefined ontology and without manual annotation?}



\begin{figure}[t]
    \centering
        \includegraphics[width=1\linewidth]{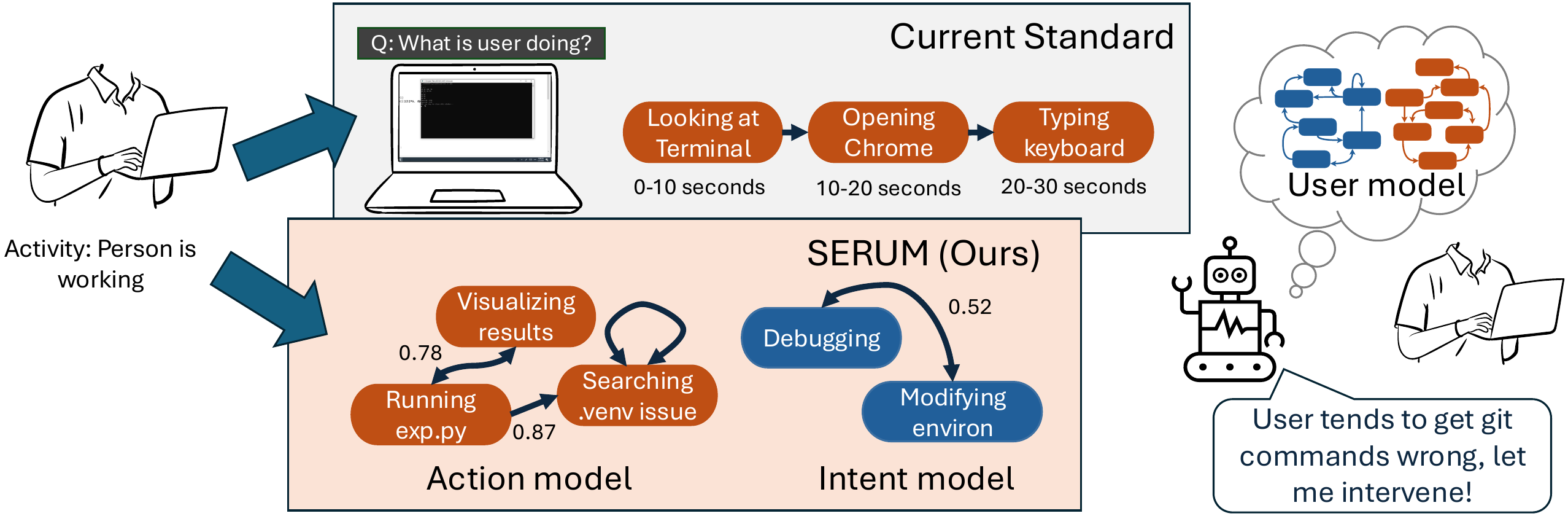}
  \caption{(Top) Current standard methods process each frame independently, producing isolated activity descriptions and coarse intent estimates. (Bottom) SERUM's multi-pass pipeline revisits prior context across frames, enabling the construction of a refined user model for both user actions and intents, enabling better informed proactive suggestions.}             
\label{fig:intro_diagram}
\end{figure}

A simple answer is to prompt a vision-language model (VLM) to label each frame, similar to recent work on general user models \cite{shaikh2025creatinggeneralusermodels}. But single-pass annotation fails in two ways: VLMs hallucinate, and they suffer from \emph{temporal conflation} — collapsing semantically distinct activities into generic labels because they lack surrounding context.

We introduce \textbf{SERUM} (\textbf{S}tate \textbf{E}xtraction and \textbf{R}efinement for \textbf{U}ser \textbf{M}odeling), a multi-pass pipeline that addresses these limitations through alternating rounds of \emph{activity recognition} and \emph{intent inference}, each grounded in the accumulated context of prior passes. SERUM operates at two complementary levels: \emph{user actions} (directly observable behaviors, e.g., pulling a git repository'') and \emph{user intents} (intermediate goals, e.g., setting up a development environment''). A sliding context window provides each pass with a run-length encoding of surrounding frames. After annotation, a \emph{label normalization} step merges synonymous labels into a compact vocabulary. SERUM outputs activity and intent models implemented as first-order Markov chains, capturing the probabilistic transition structure of user behavior.

We apply SERUM to 61 egocentric YouTube videos spanning coding, cooking, physical activity, and daily life. Our experiments show that: (1) the extracted label vocabulary reliably converges to a stable taxonomy by pass 8 — a phenomenon we term \emph{schematic equilibrium}; (2) label normalization compresses the vocabulary and sharpens transition structure, yielding better predictive models; (3) normalized Markov models outperform frequency baselines on both action and intent sequences; and (4) annotators judge final-pass labels as correct 88.3\% of the time ($\alpha$=0.40) and prefer them over first-pass labels 82.8\% of the time ($\alpha$=0.41).


SERUM is, to our knowledge, the first framework for extracting structured user activity and intent models directly from unstructured egocentric video — requiring no logs, no predefined taxonomies, and no labeled data. The resulting models represent an early validation for downstream applications such as proactive agentic assistance and personalization. Code and data are publicly available.\footnote{https://github.com/minnesotanlp/SERUM}

\section{Related Work}
\label{sec:related}

\textbf{Egocentric Video Understanding and Action Anticipation.}
Benchmarks such as EPIC-KITCHENS~\citep{damen2018epic,damen2022epickitchens100} and Ego4D~\citep{grauman2022ego4d} have established that predicting what a user will do next requires reasoning at two levels: the immediate action and the underlying goal. Mascaro et al.~\citep{mascaro2022intentionlta} exploit this hierarchy by conditioning low-level action predictions on inferred high-level intentions for long-term anticipation. \cite{furnari2020rolling} show that rolling-unrolling recurrent representations further improve anticipation on EPIC-KITCHENS. SERUM is complementary: rather than operating on labeled benchmark data, it infers both action and intent labels from scratch using raw, unannotated video.

\textbf{VLMs as Video Annotators and User Modelers.}
Foundation vision-language models such as the Qwen-VL family~\citep{bai2023qwenvl,wang2024qwen2vl,bai2025qwen3vltechnicalreport} have lowered the cost of video annotation, but single-pass inference on long videos suffers from hallucination and temporal conflation. ROVER~\citep{schroeder2025rover} and VideoNarrator~\citep{wu2025videonarrator} address these failures through recursive decomposition and multi-component verification pipelines, respectively; LLMs more broadly have been shown to match or surpass crowd workers when labels are iteratively verified~\citep{gilardi2023chatgpt,he2023annollm}. SERUM shares the insight that iterative, context-aware annotation suppresses hallucination, but goes further: each pass re-annotates frames conditioned on all prior-pass labels, enabling the model to revise earlier judgments as context accumulates and driving convergence toward a self-consistent vocabulary without any predefined ontology. Most closely related in motivation is GUM~\citep{shaikh2025creatinggeneralusermodels}, which builds user models from computer-use screenshots by inferring and revising confidence-weighted propositions about user preferences and knowledge. While GUM targets \emph{who} the user is, SERUM targets \emph{what} the user is doing and intends to do next, producing activity and intent transition models via multi-pass re-annotation rather than propositions about user traits.

\textbf{Process Mining and Behavioral Sequence Models.}
Process mining recovers structured process models from event logs~\citep{vanderaalst2012manifesto,vanderaalst2016book}, but these algorithms assume clean logs of known event types. SERUM, while conceptually related, assumes no ontology of event types and uses unstructured video as input. Process-mining quality criteria (i.e., fitness, precision, and generalization~\citep{buijs2012fitness}) motivate our use of next-action prediction accuracy and perplexity as evaluation metrics.

\textbf{Semantic Label Normalization.}
Open-vocabulary annotation produces synonymous labels that inflate state-space size and degrade model quality. We consolidate them via pairwise embedding similarity using Sentence-BERT~\citep{reimers2019sentencebert}, with a human-calibrated merging threshold ($t^{*}=0.43$). This is analogous to entity resolution and ontology alignment in knowledge-base construction. We treat normalization as a design component rather than a post-hoc fix, and empirically show it improves both vocabulary compactness and predictive accuracy of the resulting Markov models.

\section{\serum{}: \textbf{S}tate \textbf{E}xtraction and \textbf{R}efinement for \textbf{U}ser \textbf{M}odeling}
\label{sec:methodology}

\begin{figure}[t]
    \centering
    \includegraphics[width=1\linewidth]{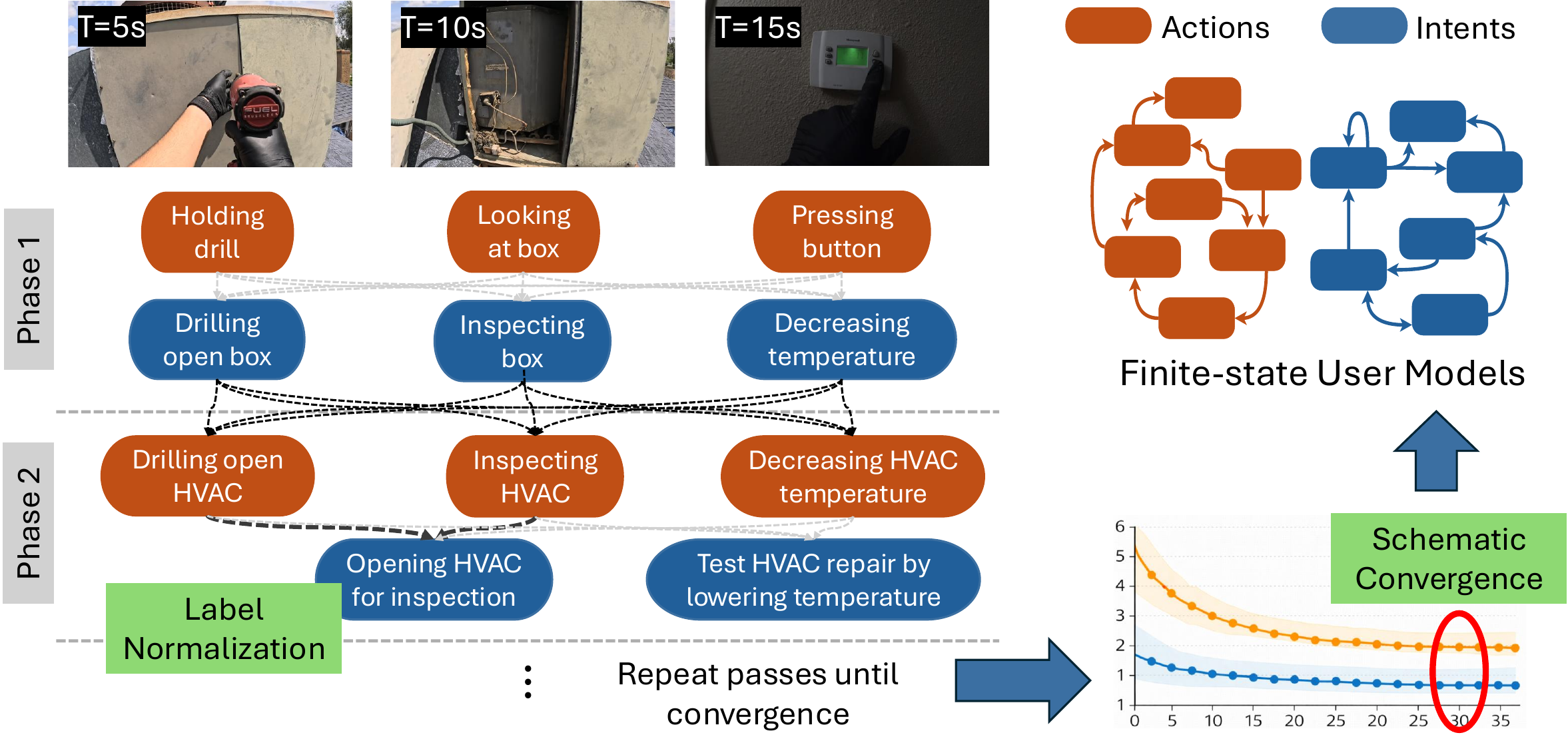}
    \caption{The \serum{} pipeline applied to an HVAC repair video. Frames are annotated through alternating activity (orange) and intent (blue) passes. Early passes yield generic labels (e.g., ``holding drill''); later passes produce fine-grained, context-aware labels (e.g., ``testing HVAC repair''). Labels are normalized before constructing the final User Models.}
    \label{fig:serum_framework}
\end{figure}

\serum{}  is a multi-pass framework for extracting structured behavioral models from raw egocentric video---no predefined ontology, no manual annotation required.
Given a sequence of $T$ sampled frames $\{f_1, \ldots, f_T\}$, \serum{} alternates between grounded activity recognition and intent inference, progressively refining coarse perceptual observations into temporally coherent behavioral descriptions.
The final output is a pair of \textbf{User Models} (UMs)---one over actions, one over intents---that compactly represent the user's behavioral dynamics and directly support downstream applications such as proactive next-action prediction.
Figure~\ref{fig:serum_framework} provides an overview of the full pipeline.

\subsection{Actions and Intents}
\label{sec:terminology}
 
\serum{} represents behavior at two distinct levels.
\textbf{Actions} ($a_t$) are mid-level natural-language descriptors of directly observable behavior at frame $t$ (e.g., ``washing vegetables,'' ``pulling a git repository'').
\textbf{Intents} ($i_t$) are latent goal-directed states, that may not be directly observable, but are inferred from sequences of actions (e.g., ``preparing dinner,'' ``setting up a dev environment'').
 
The two levels are \emph{mutually informative}: action evidence anchors intent inference; intent context disambiguates ambiguous actions.
For example, ``looking at a phone'' is labeled ``checking map directions'' once intent context establishes ``navigating to a destination.''
This bidirectionality motivates \serum{}'s alternating design---running separate independent passes for each level underperforms because neither level grounds the other. Prompt templates are provided in Appendix~\ref{appendix:prompts}.

\subsection{Multi-Pass Annotation Pipeline}
\label{sec:multipass}

A natural baseline is single-pass VLM annotation: prompt the model once per frame for both labels.
This fails for two reasons: intent inference is inherently \emph{retrospective} (the meaning of an action often only becomes clear after observing subsequent frames), and without shared context across frames, VLMs produce \emph{semantically inconsistent} labels (e.g., ``rinsing produce'' vs.\ ``cleaning vegetables'' for the same activity).
Running separate independent passes for each level does not help either. Actions and intents are \emph{mutually informative}---intent context disambiguates ambiguous actions, while action evidence anchors intent inference.
Only by \emph{alternating} the two---each pass conditioning on the outputs of the last---can they ground each other in a feedback loop converging toward coherent, disambiguated descriptions.

\serum{} implements this via alternating passes over frames $\{f_1,\ldots,f_T\}$
using \texttt{Qwen3-VL-8B-Instruct}, where odd passes annotate actions and even passes annotate intents.
Pass 1 produces unconditioned action labels $\{a_t^{(1)}\}$ in free-form natural language.
Pass 2 infers intent labels $\{i_t^{(1)}\}$ conditioned on $\{a_t^{(1)}\}$; Pass 3 refines action labels conditioned on $\{i_t^{(1)}\}$; and so on.
Each pass (from Pass 2 onward) receives two context signals.
First, a \textbf{temporal context window} of $w{=}20$ neighboring frames, encoded as a run-length encoding (RLE) that collapses consecutive identical states into count-weighted entries, providing dense local context without exceeding the model's token limit.
Second, an \textbf{inter-pass summary}---a natural-language summary generated from the full-pass RLE and the prior summary---that propagates global narrative context forward as compressed episodic memory.
For long videos exceeding $2^{15}$ tokens, a map-reduce procedure summarizes fragments independently before merging into a coherent global summary.
Passes continue until the label vocabulary stabilizes---empirically by pass~8---a convergence we term \emph{schematic equilibrium} (\S\ref{sec:rq1_convergence}).
 
\subsection{Label Normalization}
\label{sec:normalization}

Free-form annotation produces surface synonyms that inflate vocabulary size and degrade model quality.
We resolve these via Sentence-BERT~\citep{reimers2019sentencebert} cosine similarity with a human-calibrated merging threshold $t^*{=}0.43$, chosen to maximize F1 on human-judged synonym pairs.\footnote{Calibration procedure detailed in Appendix~\ref{appendix:threshold_procedure}.}
This reduces vocabulary size by 46.0\% on average and measurably improves predictive accuracy downstream.
 
\subsection{Output: User Models}
\label{sec:fusm}
\begin{figure}[t]
    \centering
    \begin{subfigure}[t]{0.60\linewidth}
        \centering
        \includegraphics[width=\linewidth]{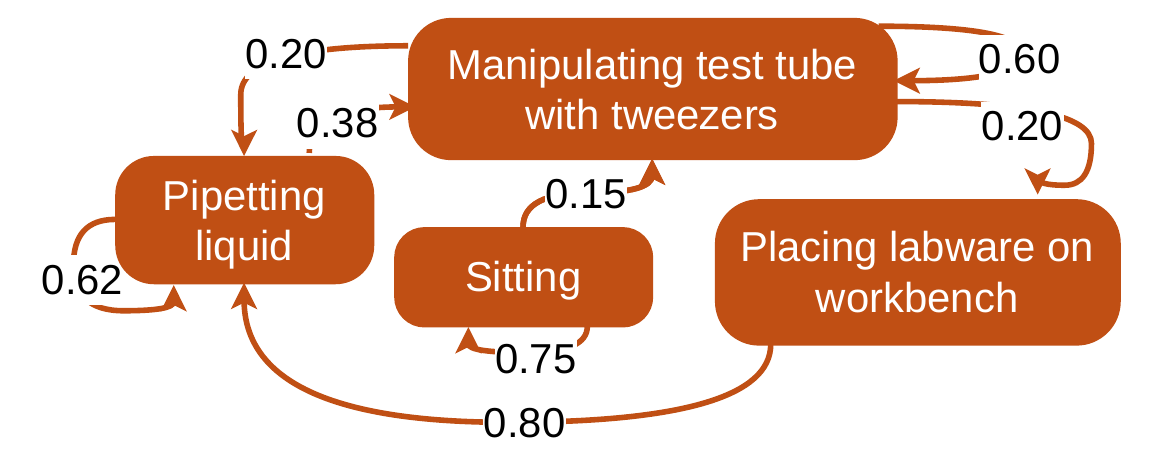}
        \caption{Action-level UM}
        \label{fig:example_activity}
    \end{subfigure}
    \hfill\hfill
    \begin{subfigure}[t]{0.37\linewidth}
        \centering
        \includegraphics[width=\linewidth]{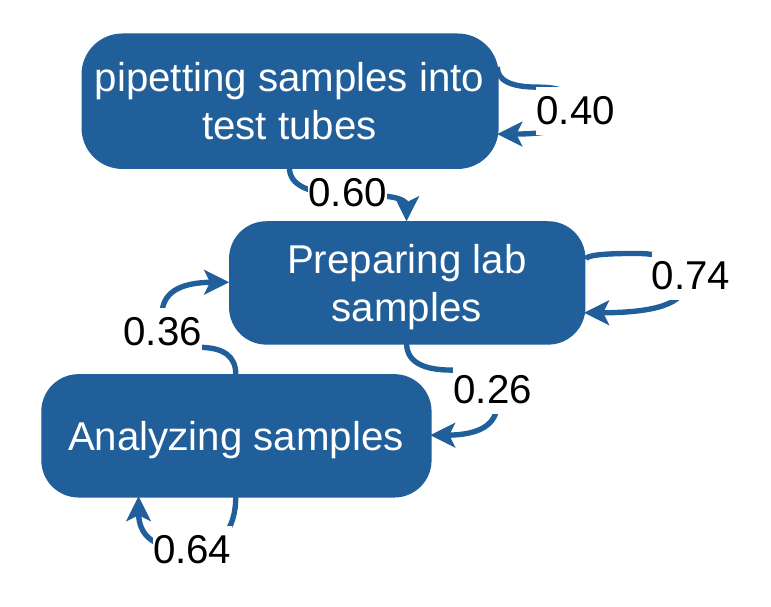}
        \caption{Intent-level UM}
        \label{fig:example_intent}
    \end{subfigure}
    \caption{Example User Models from a single video. Action-level states (a) capture observable behaviors; intent-level states (b) capture inferred goals.}
    \label{fig:example_fusm}
\end{figure}

 The final output is a pair of \textbf{User Models} (UMs): directed weighted graphs $\mathcal{M}{=}(\mathcal{S},\mathcal{E})$ where $\mathcal{S}$ is the canonicalized state vocabulary and each edge $(s,s')$ is weighted by observed transition frequency.
One UM is built over actions, one over intents, yielding complementary views of behavior (Figure~\ref{fig:example_fusm}).
 UMs support proactive assistance by surfacing probable next states given the user's current state.
We evaluate predictive utility via a \textbf{next-action prediction} task: UMs trained on the first 60\% of frames predict the held-out final 40\%, measuring whether captured behavioral dynamics generalize to unseen activity.

\section{Evaluation}

We evaluate \serum on 61 egocentric videos spanning coding, cooking, physical activity, and daily life, addressing four research questions:
\textbf{RQ1}~(\S\ref{sec:rq1_convergence}) does \serum's iterative annotation converge
to a stable label vocabulary (\emph{schematic equilibrium})?
\textbf{RQ2}~(\S\ref{sec:rq2_prediction}) do the resulting user models usefully predict
next user states?
\textbf{RQ3}~(\S\ref{sec:rq3_human}) are \serum's labels aligned with human judgment?
\textbf{RQ4}~(\S\ref{sec:rq4_ablation}) how do label normalization and intent passes
each contribute to model quality?

\subsection{Experimental Setup}
\textbf{Dataset}\label{sec:dataset}
We evaluate on 61 egocentric videos sampled at 5-second intervals, yielding 11,125 total
frames across four domains (Table~\ref{tab:domain-stats}).

\begin{table}[t]
\centering
\caption{Per-domain dataset statistics. Vocabulary and accuracy from the final activity (P11) and intent (P12) passes. Values are mean $\pm$ std. Act.\ = Activity, Int.\ = Intent. We find that coding videos tend to consist of repetitive actions (writing code) focused on a singular goal (releasing a snake game), resulting in higher accuracy.}
\label{tab:domain-stats}
\resizebox{\textwidth}{!}{
\begin{tabular}{lrrrrrr}
\toprule
Domain & Videos & Frames & Act.\ Vocab & Int.\ Vocab & Act.\ Acc (\%) & Int.\ Acc (\%) \\
\midrule
Coding & 19 & $207 \pm 85$ & $7 \pm 3$ & $26 \pm 22$ & $73.9$ & $47.5$ \\
Cooking & 15 & $219 \pm 101$ & $66 \pm 29$ & $48 \pm 24$ & $13.6$ & $21.8$ \\
Physical & 12 & $155 \pm 81$ & $29 \pm 17$ & $24 \pm 11$ & $25.7$ & $40.1$ \\
Daily Life & 15 & $136 \pm 81$ & $28 \pm 21$ & $39 \pm 26$ & $24.7$ & $21.5$ \\
\midrule
\textbf{All} & \textbf{61} & \textbf{$182 \pm 94$} & \textbf{$31 \pm 29$} & \textbf{$34 \pm 24$} & \textbf{$37.5$} & \textbf{$33.3$} \\
\bottomrule
\end{tabular}}  
\\[2pt]\footnotesize 61 videos, 11,125 frames, 927 min (15.5 hrs), 12 passes each, 133,500 total state extractions. A generalization study on EPIC-KITCHENS-100 (366 videos, 37 participants) is reported in Appendix~\ref{appendix:appendix_generalization}.             
\end{table}

\textbf{Model and inference.}
All annotation passes use Qwen3-VL-8B-Instruct~\citep{bai2025qwen3vltechnicalreport} \footnote{Preliminary study on model choice in Appendix~\ref{appendix:appendix_model_choice}} in BF16, served via vLLM~\citep{kwon2023vllm} with tensor parallelism across two GPUs per node. We distribute inference across two nodes (2$\times$ NVIDIA A5000, 2$\times$ NVIDIA A6000), each running an independent vLLM server, achieving a combined throughput of 1.3 inferences/sec and processing 12 passes for a 10-minute video in $\approx$17 minutes per node.

\textbf{Label Normalization.}
We apply pairwise semantic merging using SentenceBERT
embeddings~\citep{reimers2019sentencebertsentenceembeddingsusing} with cosine-distance threshold $t^{*} = 0.43$,
selected to maximize F1 on a human-annotated calibration set of 100 activities and 100 intents randomly sampled from all passes.
(\S\ref{tab:canon_summary}).

\subsection{RQ1: Schematic Equilibrium}\label{sec:rq1_convergence}

\textbf{Setup.}
To determine how many annotation passes are needed, we ran a pilot study on 13 videos
for 30 passes, tracking vocabulary size per pass before and after label
normalization.
 
\begin{figure}[H]
    \centering
    \begin{minipage}[c]{0.3\linewidth}
        \begin{subfigure}[b]{\linewidth}
            \includegraphics[width=0.9\linewidth]{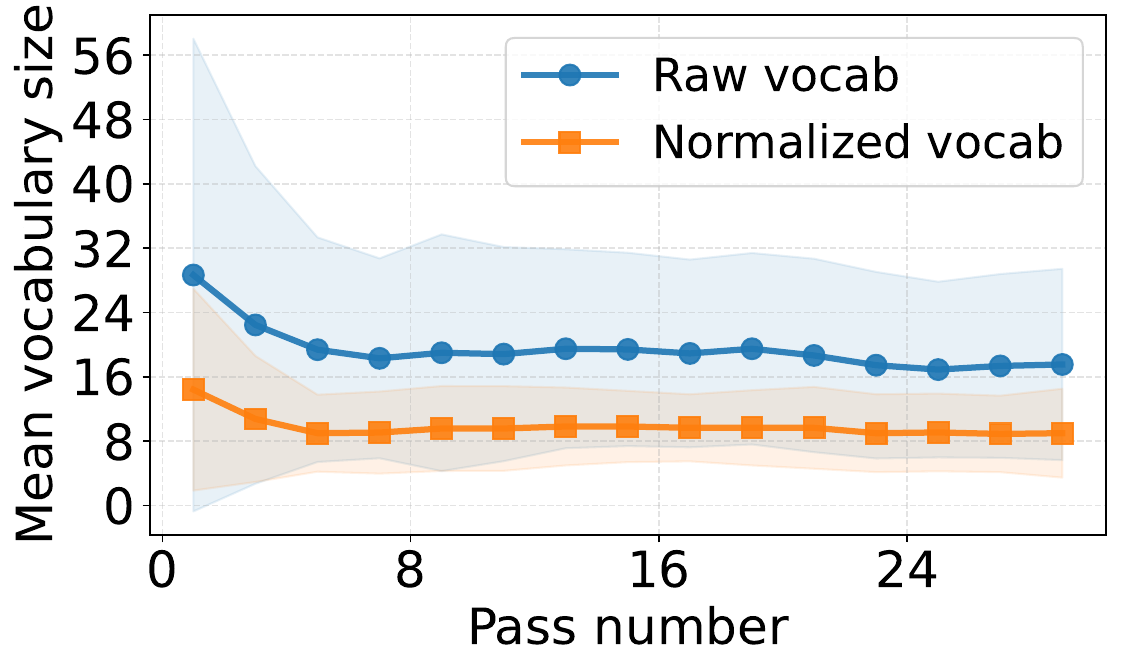}\vspace{-2mm}
            \caption{Action: raw vs.\ normalized}
            \label{fig:pilot_canon_activity}
        \end{subfigure}
        \vfill
        \begin{subfigure}[b]{\linewidth}
            \includegraphics[width=0.9\linewidth]{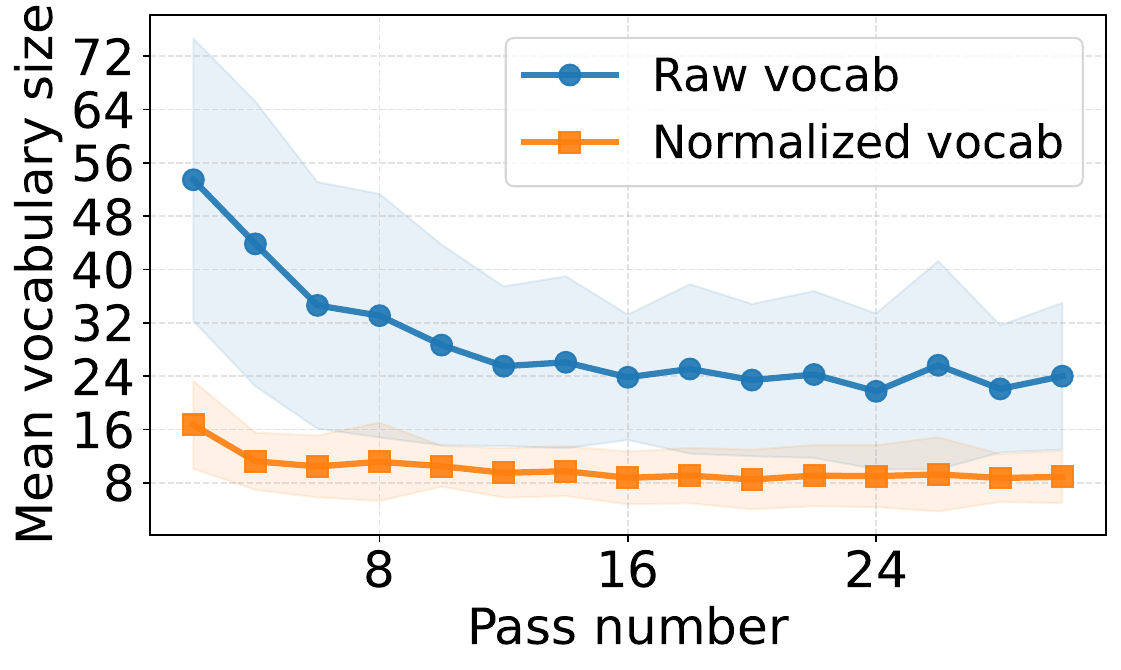}\vspace{-2mm}
            \caption{Intent: raw vs.\ normalized}
            \label{fig:pilot_canon_intent}
        \end{subfigure}
    \end{minipage}
    \hfill
    \begin{minipage}[c]{0.65\linewidth}
        \begin{subfigure}[b]{\linewidth}
            \includegraphics[width=\linewidth]{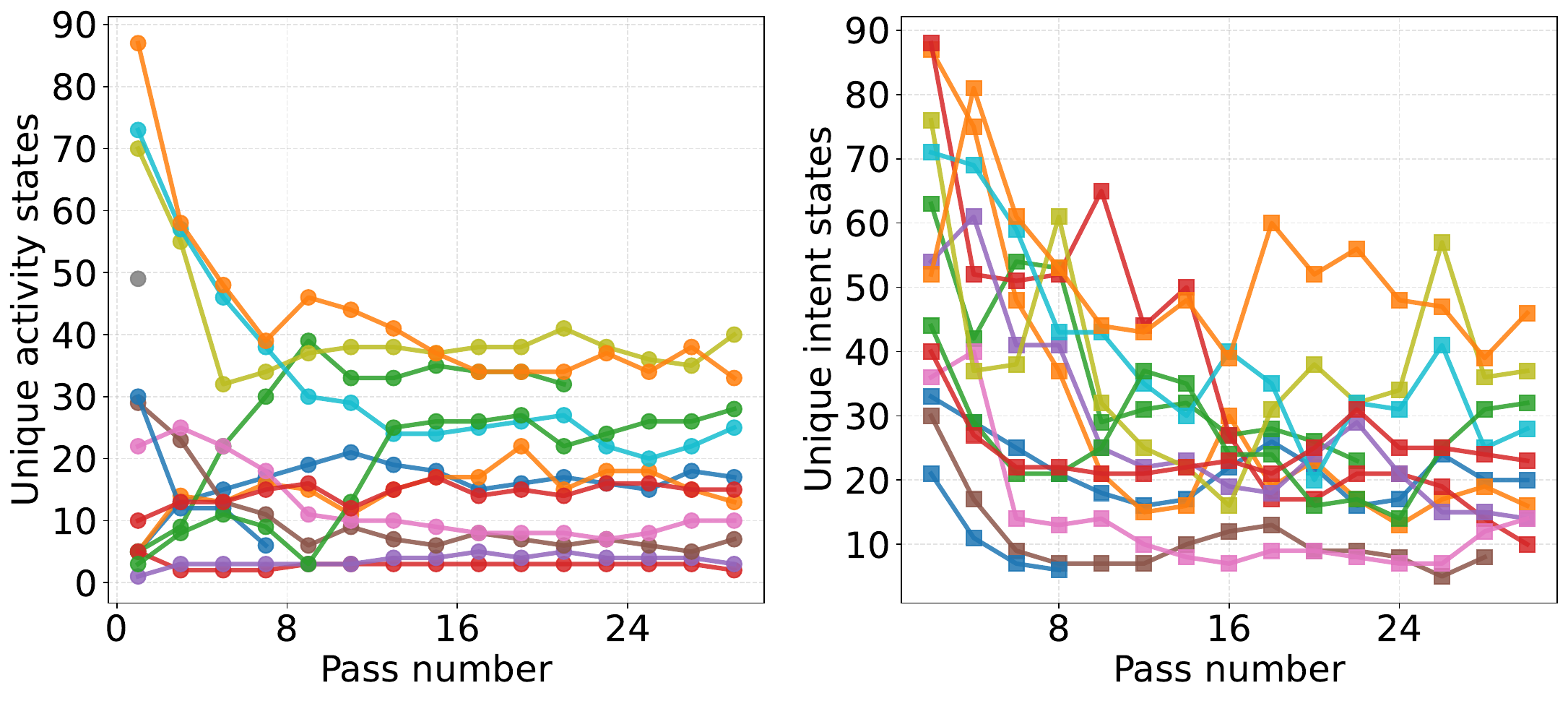}
            \caption{Per-video vocabulary across passes. Left: Activity, Right: Intent}
            \label{fig:pilot_vocab}
        \end{subfigure}
    \end{minipage}
    \caption{30-pass pilot study (13 videos). (a) Activity vocabulary drops from~28 to~18 raw
    states by pass~8, with normalization compressing further to~10. (b) Intent vocabulary shows
    a steeper decline from~54 to~24 raw states, with normalization consistently reducing to~10
    across all passes. (c) Per-video vocabulary stabilizes by pass~8 (\emph{schematic equilibrium}).}
    \label{fig:pilot_convergence}
\end{figure}

\textbf{Results.}
As shown in Figure~\ref{fig:pilot_convergence}, the average raw activity vocabulary drops from
$\sim$28 to $\sim$18 unique states by pass~8, while the average raw intent vocabulary drops more
steeply from $\sim$54 to $\sim$24. Label normalization compresses both further to
$\sim$10 states and remains stable thereafter. Per-video vocabulary curves
(Figure~\ref{fig:pilot_convergence}c) confirm that all videos individually stabilize by
pass~8 — a convergence we term \emph{schematic equilibrium}. Based on this finding, we
run all large-scale evaluations at 12 passes (6 activity and 6 intent, interleaved),
providing a margin beyond the observed convergence point.

\subsection{RQ2: Next-State Prediction}
\label{sec:rq2_prediction}
\textbf{Setup.}
We construct Markov user models from the first 60\% of frames per video and evaluate on
the remaining 40\%, applying add-one Laplace smoothing to transition counts. We compare
against three baselines: \textbf{Majority} (always predict the most frequent training
state), \textbf{Weighted Random} (sample proportional to marginal frequency), and
\textbf{Uniform} (sample uniformly over observed states). Performance is measured by
\textbf{top-1 accuracy} and \textbf{perplexity} (exponentiated cross-entropy over
held-out transitions; lower is better).

\begin{table}[t]
\centering
\small
\caption{Mean top-1 accuracy and perplexity at the final pass.
$^{n}$ denotes normalized labels.}
\label{tab:cross_video_accuracy}
\setlength{\tabcolsep}{5pt}
\begin{tabular}{lrrrr}
\toprule
& \multicolumn{2}{c}{Activity} & \multicolumn{2}{c}{Intent} \\
\cmidrule(lr){2-3}\cmidrule(lr){4-5}
Model & Top-1~($\uparrow$) & PPL~($\downarrow$) & Top-1~($\uparrow$) & PPL~($\downarrow$) \\
\midrule
Markov        & 37.5{\small$\pm$33.3} & 23.9{\small$\pm$25.6} & 33.3{\small$\pm$26.9} & 24.6{\small$\pm$21.5} \\
Majority      & 37.6{\small$\pm$34.0} & 29.8{\small$\pm$32.0} & 29.9{\small$\pm$28.1} & 31.8{\small$\pm$35.5} \\
Wt.\ Random   & 25.9{\small$\pm$29.5} & 29.8{\small$\pm$32.0} & 17.9{\small$\pm$20.4} & 31.8{\small$\pm$35.5} \\
Uniform       &  9.2{\small$\pm$10.5} & 30.8{\small$\pm$29.0} &  6.5{\small$\pm$ 9.0} & 34.2{\small$\pm$24.3} \\
\midrule
Markov$^{n}$  & \textbf{48.5}{\small$\pm$30.9} & \textbf{10.9}{\small$\pm$11.0}
              & \textbf{58.2}{\small$\pm$30.3} & \textbf{8.2}{\small$\pm$ 9.7} \\
Majority$^{n}$& 46.5{\small$\pm$32.6} & 14.4{\small$\pm$15.8} & 53.4{\small$\pm$33.8} & 11.0{\small$\pm$14.7} \\
\bottomrule
\end{tabular}
\end{table}
 
\textbf{Main Result.} Table~\ref{tab:cross_video_accuracy} shows that at the final annotation pass, Markov user models outperform naive baselines in both top-1 accuracy and perplexity.\footnote{Top-3 and Top-5 show similar tendencies but more strongly favor Markov.} 
Normalized variants (superscript $n$) apply post-hoc label normalization merging prior to model construction. Normalization benefits the Markov model disproportionately, since fewer labels reduces sparsity in its transition matrix, whereas the Majority baseline only tracks label frequencies.

\begin{table}[h]
\centering
\small
\caption{Markov top-1 accuracy and perplexity by domain (final pass, normalized labels).}
\label{tab:domain-results}
\setlength{\tabcolsep}{4pt}
\begin{tabular}{lrrrrrrr}
\toprule
& & \multicolumn{2}{c}{Act Top-1 (\%)} & \multicolumn{2}{c}{Intent Top-1 (\%)}
& \multicolumn{2}{c}{Perplexity ($\downarrow$)} \\
\cmidrule(lr){3-4}\cmidrule(lr){5-6}\cmidrule(lr){7-8}
Domain & $n$ & Raw & Norm. & Raw & Norm. & Activity & Intent \\
\midrule
Coding & 19 & 73.9 & 76.4 & 47.5 & 76.1 &  2.4 & 2.9 \\
Cooking       & 15 & 13.6 & 31.8 & 21.8 & 53.8 & 16.0 & 6.7 \\
Physical      & 12 & 25.7 & 44.2 & 40.1 & 71.6 & 11.2 & 4.1 \\
Daily Life    & 15 & 24.7 & 33.5 & 21.5 & 30.3 & 14.2 & 17.5 \\
\midrule
\textbf{Overall} & \textbf{61}
  & \textbf{37.5} & \textbf{48.5}
  & \textbf{33.3} & \textbf{58.2}
  & \textbf{23.9} & \textbf{24.6} \\
\bottomrule
\end{tabular}
\end{table}
 

Domain-level results (Table~\ref{tab:domain-results}) show largest gains on
structured tasks: Coding achieves 76.4\% normalized activity accuracy,
reflecting the rich, repetitive transition structure of coding workflows. Cooking and
physical tasks see smaller absolute accuracy but substantial relative gains from
normalization. We further find preliminary evidence that SERUM-produced markov models can transfer to similar but unseen workflow videos. \footnote{Preliminary study detailed in Appendix~\ref{appendix:transferability_study}.}


\subsection{RQ3: Human Assessment of Label Quality}
\label{sec:rq3_human}
\textbf{Setup.}
We recruited five colleagues with domain expertise in human workflow research to
evaluate label quality. Annotators assessed 180 uniformly sampled frames from 9
videos (10 activity, 10 intent per video), of a 26 video pre-vetted set \footnote{See the Ethics Statement for pre-vetting criteria.}, presented via a web application embedding the source
video at the relevant timestamp. For each frame, annotators judged: (1)~\textbf{label
accuracy} — whether the final-pass label correctly describes the observed activity or
intent; and (2)~\textbf{pass preference} — whether the final-pass or first-pass label is
better. Responses are aggregated by majority vote; inter-annotator agreement (IAA) is
measured via Krippendorff's $\alpha$.

\begin{figure}[h]
\centering
\small
\caption{Human annotation results and error analysis}
\label{fig:human-annotation}

\begin{minipage}[c]{0.48\linewidth}
\centering

\begin{tabular}{lrrrr}
\toprule
Annotator & Acc. & Inacc. & Final & First \\
\midrule
A        & 88.3 & 11.7 & 78.9 & 21.1 \\
B        & 79.4 & 20.6 & 77.2 & 22.8 \\
C        & 76.1 & 23.9 & 76.1 & 23.9 \\
D        & 86.7 & 13.3 & 80.0 & 20.0 \\
E        & 92.8 & 7.2  & 82.8 & 17.2 \\
\midrule
Majority vote & 88.3 & 11.7 & 82.8 & 17.2 \\
\bottomrule
\end{tabular}

\smallskip
{\small (a) Per-annotator accuracy and pass preference.}

\vspace{8pt}

\begin{tabular}{lrrrr}
\toprule
Type & $n$ & $\alpha_{\text{acc}}$ & Maj.\ Acc & Final Pref \\
\midrule
Activity & 90  & 0.399 & 86.7\% & 82.8\% \\
Intent   & 90  & 0.402 & 90.0\% & 81.1\% \\
\bottomrule
\end{tabular}

\smallskip
{\small (b) Inter-annotator agreement by label type.}

\end{minipage}
\hfill
\begin{minipage}[c]{0.50\linewidth}
\centering
\includegraphics[width=\linewidth]{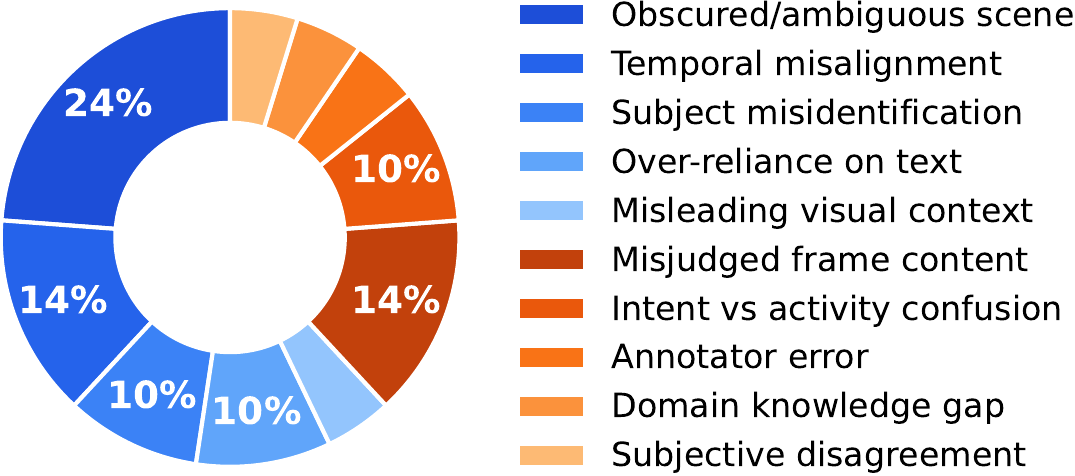}

\smallskip
{\small (c) Distribution of inaccuracy causes, VLM (blue) was responsible for 62\% of issues, annotator labeling (orange) caused 38\% of misjudgements}
\end{minipage}

\end{figure}

\textbf{Results.}
By majority vote, \textbf{88.3\%} of labels were rated accurate ($\alpha = 0.40$) and
final-pass labels were preferred over first-pass labels \textbf{82.8\%} of the time
($\alpha = 0.41$), suggesting that iterative annotation produces meaningfully better
labels (Figure~\ref{fig:human-annotation}). Intent labels
were slightly more accurate than Activity labels (90\% vs.\ 86.7\%), yet preference rates
were comparable across both types (81.1\% vs.\ 82.8\%), indicating that multi-pass
refinement improves intent inference similarly to activity recognition despite intent
being harder to verify from a single frame.

\subsection{RQ4: Ablation Study}
\label{sec:rq4_ablation}

\textbf{Setup.}
We isolate the contribution of the three core design choices in \serum.
For \textbf{label norm}, we compare Markov models built on raw vs.\ normalized
label sequences at the final pass, measuring vocabulary reduction, top-1 accuracy, and
perplexity. For \textbf{intent passes}, We compare the full \serum pipeline against an activity-only baseline pipeline in three scenarios. (1) both pipelines infer and evaluate on their own freely generated vocabularies, (2) both pipelines initially infer on open vocabularies, but project activity-only labels on to the intent-conditioned vocabulary before evaluation, (3) Project in reverse direction. For \textbf{temporal window size}, we compare SERUM's default temporal window of n=20 with n=10 and n=0 on accuracy and perplexity against respective baselines.

\begin{figure}[h]
    \centering\vspace{-4mm}
    \begin{subfigure}[c]{0.40\linewidth}
        \centering
        \small
        \caption{Effect of label normalization on vocabulary and Markov model quality.}
        \label{tab:canon_summary}
        \begin{tabular}{@{}lr@{}}
        \toprule
        Metric & Value \\
        \midrule
        Vocab.\ reduction         & $46.0\%\ \pm\ 21.1\%$ \\
        Top-1 acc change     & $+18.2\ \pm\ 21.2\,\text{pp}$ \\
        Perplexity change         & $-14.90\ \pm\ 17.19$ \\
        \bottomrule
        \end{tabular}
    \end{subfigure}
    \hfill
    \begin{subfigure}[c]{0.27\linewidth}
        \centering
        \small
        \begin{tabular}{@{}lr@{}}
        \toprule
        Metric & Value \\
        \midrule
        Optimal $t^{*}$      & 0.43 \\
        F1                   & 0.8217 \\
        Precision            & 0.7681 \\
        Recall               & 0.8833 \\
        Same pairs           & 60     \\
        Different pairs      & 140    \\
        Mean same dist.      & 0.2471 \\
        Mean diff.\ dist.    & 0.7864 \\
        \bottomrule
        \end{tabular}
        \caption{Calibration statistics}
        \label{fig:threshold_table}
    \end{subfigure}
    \hfill
    \begin{subfigure}[c]{0.27\linewidth}
        \centering
        \includegraphics[width=\linewidth]{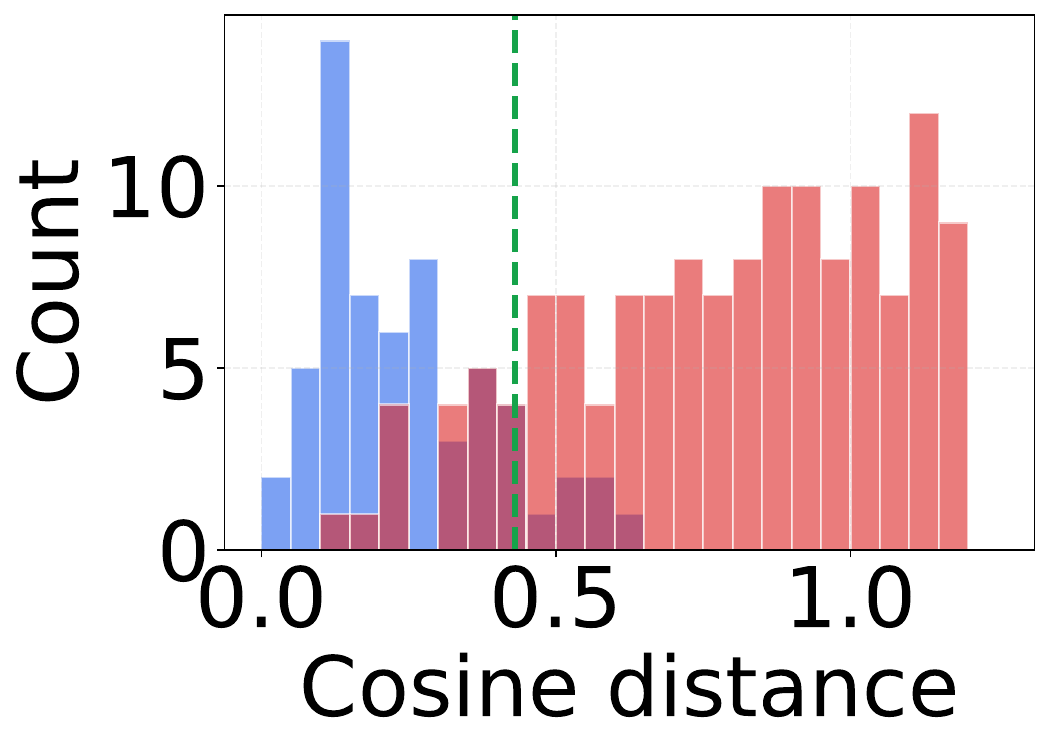}
        \caption{Distribution for synonymous (blue) and
        distinct (red) pairs. Dashed line marks $t^{*}\!=\!0.43$ }
        \label{fig:threshold_plot}
    \end{subfigure}
    \caption{Label normalization summary (left) and semantic threshold calibration on 200 human-annotated label pairs (right).}
    \label{fig:threshold_calibration}
\end{figure}

\textbf{Label normalization.}
Normalization reduces the state vocabulary by \textbf{46.0\%} on average
($\pm$21.1\%) while improving Markov top-1 accuracy by \textbf{$+$18.2\,pp}
($\pm$21.2\,pp) and reducing perplexity by 14.9 points 
Gains are consistent across domains and largest for intent models, where surface-synonym
proliferation is most severe. This validates open-vocabulary annotation followed by
principled merging as better than a fixed ontology: the former preserves
fine-grained behavioral distinctions that the latter would collapse, and normalization then recovers the compact transition structure needed for reliable Markov estimation.
The calibrated SentenceBERT threshold ($t^{*}\!=\!0.43$, F1\,=\,0.822;
Figure~\ref{fig:threshold_table} and \ref{fig:threshold_plot}) separates synonymous from distinct labels with high precision (0.768) and recall (0.883).

\textbf{Value of intent passes.}
To isolate the contribution of intent gathering annotation passes, we compare the full pipeline against an activity-only baseline using 12 activity passes with no intent inference. The preference study below tests whether this predictability reflects genuine label quality. Across 61 videos, 47\% of frames received different activity labels between the two conditions. We sampled 30 of these divergent frames (10 per domain, stratified across 3 videos) and presented each as a blinded A/B pair.

By majority vote, annotators preferred labels from the full pipeline
\textbf{73\%} of the time ($\alpha = 0.726$; Table~\ref{tab:ablation-preference}).
Without intent context, activity labels collapse to uninformative dominant states:
\texttt{typing\_on\_keyboard} for 92--96\% of coding domain frames (vs.\ the
intent-informed pipeline's \texttt{editing\_css\_style},
\texttt{editing\_html\_code}, \texttt{debugging\_code}).

\begin{table}[t]
\centering
\small
\caption{Ablation preference study: full pipeline (activity+intent) vs.\ activity-only labels.
Three annotators evaluated 30 blinded A/B pairs across three domains.}
\label{tab:ablation-preference}
\begin{tabular}{lrrr}
\toprule
Annotator & $n$ & Full Pref.\ (\%) & Act.-Only Pref.\ (\%) \\
\midrule
A & 30 & 80 & 20 \\
B & 30 & 73 & 27 \\
C & 30 & 63 & 37 \\
\midrule
Majority & 30 & \textbf{73} & 27 \\
\bottomrule
\end{tabular}
\hspace{1em}
\begin{tabular}{lr}
\toprule
Metric & Value \\
\midrule
Krippendorff's $\alpha$ & 0.726 \\
A vs B agree & 93\% \\
A vs C agree & 83\% \\
B vs C agree & 90\% \\
\bottomrule
\end{tabular}
\end{table}

\textbf{Value of intent passes with frozen vocabularies.}
To study the contribution of intent passes, notwithstanding differences in vocabulary size produced by the activity-only and the full pipelines, we project the vocabulary produced by the full pipeline onto the activity-only pipeline vocabulary, project the activity-only pipeline onto the full pipeline vocabulary, and test the respective performances of both vocabularies. Both normalizations support the conclusion that vocabulary size differences do not significantly impact results.

\begin{table}[H]\centering\small
\caption{full vs.\ activity-only vocabulary before and after projection. OOV measures percent of labels that have no match in target vocabulary.}
\label{tab:exp1}
\begin{tabular}{lrrrrr}
\toprule
Condition & Vocab & Markov & Majority & PPL & OOV \\
\midrule
\multicolumn{6}{l}{\textit{Raw (own vocabulary)}} \\
Intent-conditioned & $29.9{\pm}27.4$ & $36.4{\pm}30.8$ & $36.8{\pm}32.1$ & $22.5{\pm}23.1$ & --- \\
Activity-only & $28.0{\pm}26.8$ & $42.6{\pm}34.0$ & $43.6{\pm}34.2$ & $20.7{\pm}21.7$ & --- \\
\midrule
\multicolumn{6}{l}{\textit{Normalized $\to$ intent vocab}} \\
Intent-conditioned & $18.4{\pm}14.8$ & $45.9{\pm}28.9$ & $42.6{\pm}30.4$ & $11.9{\pm}10.9$ & 1.2 \\
Activity-only & $17.2{\pm}14.5$ & $50.2{\pm}31.0$ & $46.5{\pm}33.5$ & $11.1{\pm}10.8$ & 2.0 \\
\midrule
\multicolumn{6}{l}{\textit{Normalized $\to$ activity-only vocab}} \\
Intent-conditioned & $16.9{\pm}13.6$ & $48.4{\pm}28.0$ & $44.6{\pm}29.9$ & $10.6{\pm}10.0$ & 7.2 \\
Activity-only & $17.8{\pm}15.2$ & $50.0{\pm}31.2$ & $46.8{\pm}33.3$ & $11.3{\pm}11.5$ & 1.2 \\
\bottomrule
\end{tabular}
\end{table}

\textbf{Effect of temporal window size.}
To study the effects of various temporal window sizes, we evaluate over the same 12 randomly chosen videos at window sizes n = \{0,10,20\}. Markov - Majority gap increases at higher window size (4.4 vs 8.0), suggesting prediction structures become more prominent in produced Markov models at higher window sizes.

\begin{table}[H]\centering\small
\caption{Temporal-window sensitivity ($w \in \{0, 10, 20\}$), $n=12$ videos. Markov / Majority in \%. Normalized: all conditions projected onto the $w=20$ vocabulary.}
\label{tab:exp2}
\begin{tabular}{lrrrr}
\toprule
Condition & Avg. vocab & Markov & Majority & Perplexity \\
\midrule
\multicolumn{5}{l}{\textit{Raw vocab}} \\
$w=0$ (no temporal context) & $63.00 \pm 32.74$ & $19.2 \pm 25.3$ & $19.3 \pm 25.5$ & $47.76 \pm 24.55$ \\
$w=10$                       & $55.08 \pm 29.10$ & $19.8 \pm 23.9$ & $16.6 \pm 25.9$ & $41.51 \pm 20.32$ \\
$w=20$ (paper default)       & $54.50 \pm 26.94$ & $21.5 \pm 23.8$ & $19.4 \pm 25.3$ & $40.53 \pm 19.64$ \\
\midrule
\multicolumn{5}{l}{\textit{Normalized}} \\
$w=0$                        & $30.08 \pm 14.63$ & $28.4 \pm 24.3$ & $24.0 \pm 25.8$ & $20.38 \pm 11.86$ \\
$w=10$                       & $27.92 \pm 13.69$ & $28.0 \pm 24.9$ & $21.7 \pm 27.0$ & $18.98 \pm 11.61$ \\
$w=20$                       & $30.75 \pm 14.68$ & $32.6 \pm 27.3$ & $24.6 \pm 25.5$ & $20.46 \pm 11.97$ \\
\bottomrule
\end{tabular}
\end{table}

\begin{figure}[H]
\label{fig:passwise_markov_coding}
    \centering
    \begin{subfigure}[b]{0.37\linewidth}
        \includegraphics[width=\linewidth]{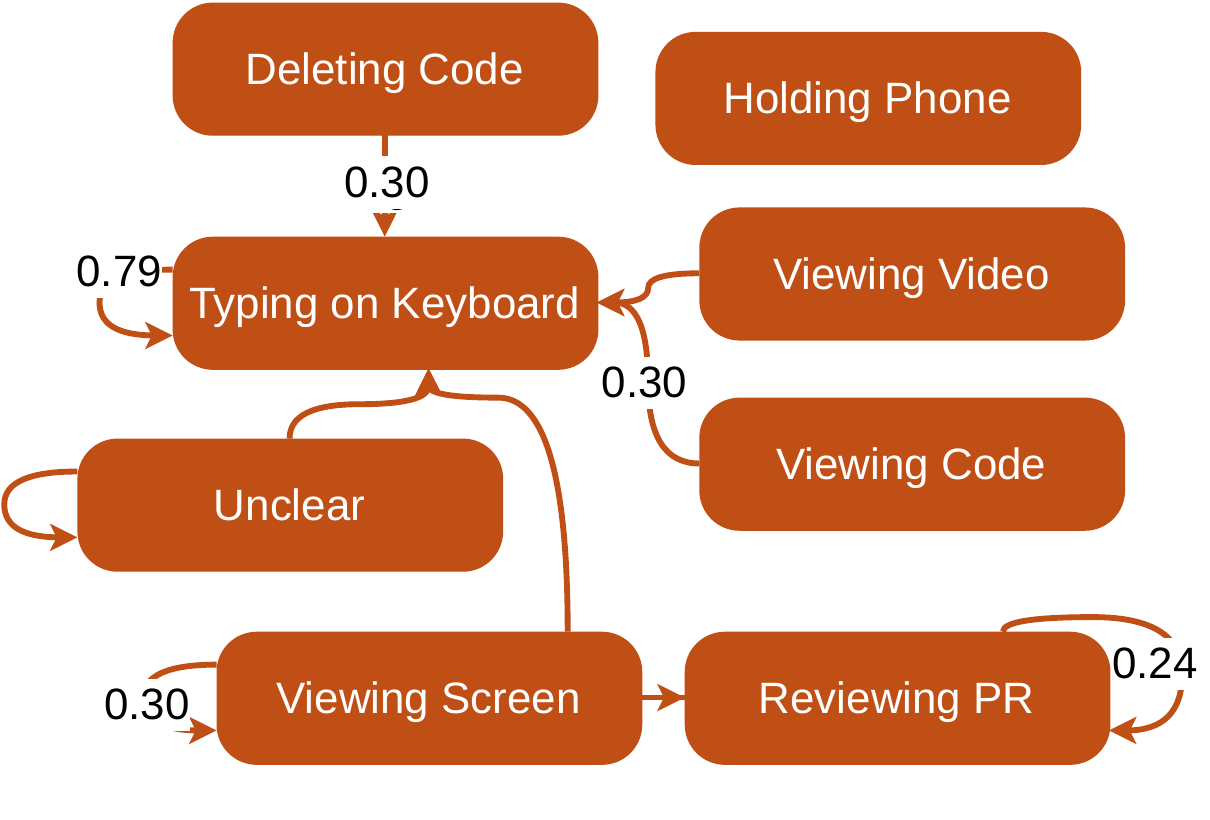}
        \caption{Pass 1 (Activity)}
    \end{subfigure}
    \hfill
    \begin{subfigure}[b]{0.37\linewidth}
        \includegraphics[width=\linewidth]{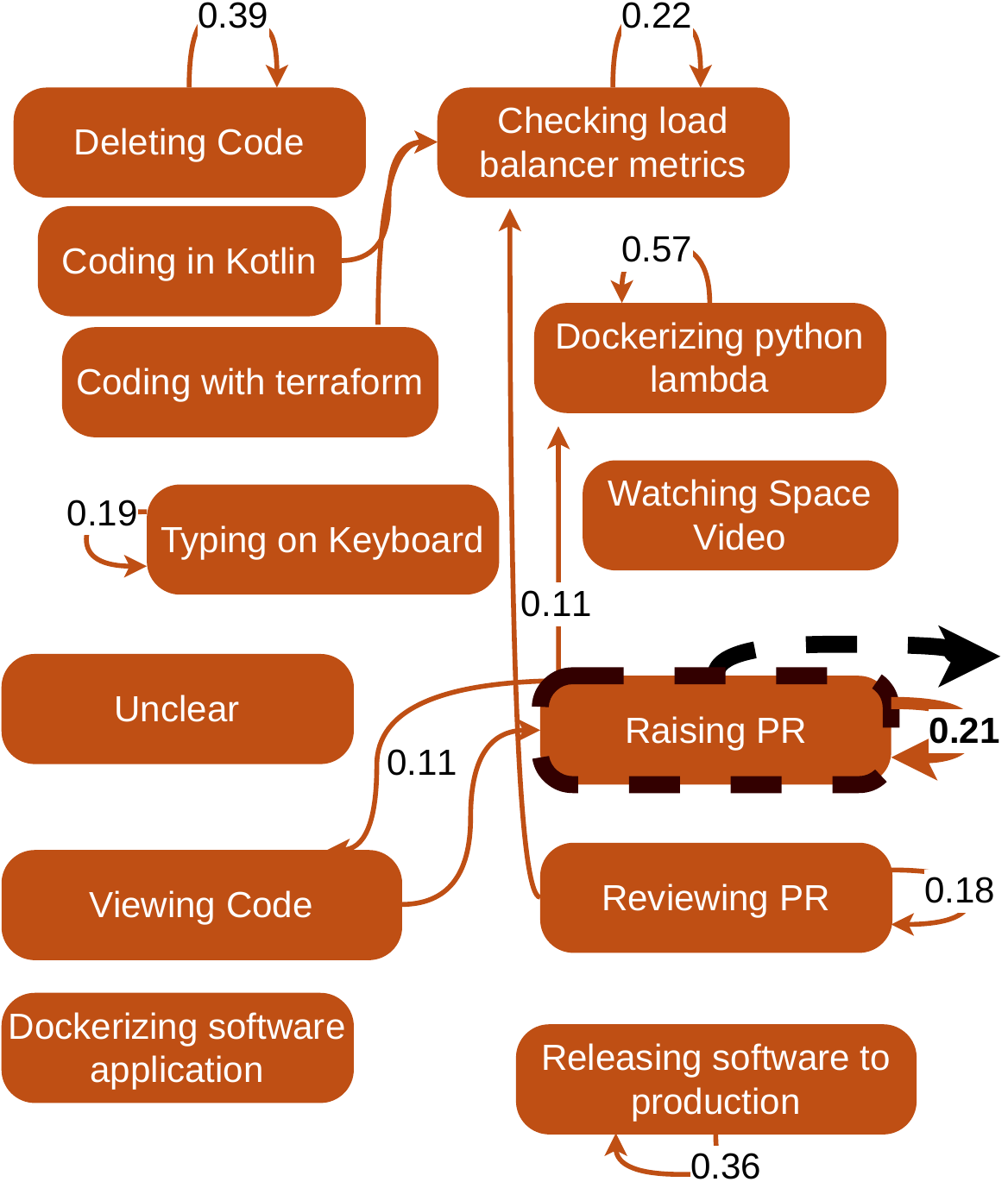}
        \caption{Pass 11 (Activity)}
    \end{subfigure}
    \hfill
    \begin{subfigure}[b]{0.24\linewidth}
        \centering
        \small
        \setlength{\tabcolsep}{0pt}
        \includegraphics[width=0.75\linewidth]{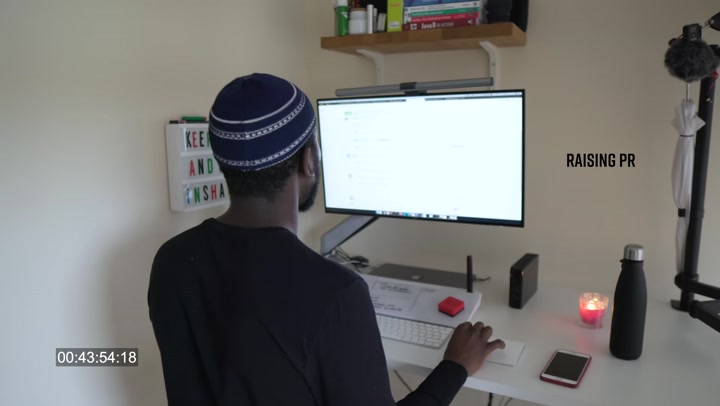}\\[4pt]
        \textbf{Activity:} \texttt{raising PR}\\
        \textbf{Next Act:} \texttt{raising PR}\\[4pt]
        \begin{tabular}{@{}lr@{}}
        \toprule
        \textbf{Markov top-3} \\
        \midrule
        \texttt{\textbf{raising PR}} & \textbf{21\% $\checkmark$} \\
        \texttt{dockerizing...} & 11\% \\
        \texttt{viewing code} & 11\% \\
        \midrule
        \textbf{Majority top-3} & \\
        \midrule
        \texttt{dockerizing...} & 31\% \\
        \texttt{deleting code} & 12\% \\
        \texttt{releasing s...} & 11\% \\
        \bottomrule
        \end{tabular}
        \caption{at $t{=}7$:05}
    \end{subfigure}
    \caption{Activity refinement and next-state prediction for \texttt{behindP12}. (a) Pass~1 produces generic labels. (b) By pass~11, task-specific states emerge. (c) The Markov model correctly predicts state persistence by conditioning on the current state, while the majority baseline erroneously predicts the three globally most frequent states regardless of context.}
    \label{fig:qualitative_prediction}
\end{figure}

\textbf{Qualitative Study.}
Figure~\ref{fig:qualitative_prediction} shows multi-pass refinement and its predictive consequence on a daily life video.\footnote{Additional examples in Appendix, Figure~\ref{fig:passwise_markov_tutorial}.}

\section{Conclusion and Discussion}
\label{sec:conclusion}
 
We presented \serum, a multi-pass VLM framework that extracts structured activity and
intent models from egocentric video without a predefined ontology or manual annotation.
Alternating activity and intent passes converge to a stable vocabulary (\emph{schematic
equilibrium}) by pass~8; subsequent label normalization compresses it by 46\%, yielding
Markov user models that outperform frequency baselines on next-state prediction.
Human annotators rate 88.3\% of final-pass labels accurate and prefer them over first-pass
labels 82.8\% of the time, suggesting that iterative refinement produces meaningful,
recognizable improvements.
This work has the following limitations and interesting directions for future work:

\textbf{Evaluation protocol.}
Split-half evaluation penalizes Markov models on videos whose content progresses linearly without revisiting earlier states, since training and test vocabularies become largely disjoint. This effect can be seen with several videos achieving near-zero accuracy before normalization (\S\ref{tab:markov-final}). Future work could address this through cross-video evaluation, where models trained on one user's videos predict states in another's.

\textbf{Downstream applications.}
An important open question is whether \serum's user models can drive proactive
agentic assistance — anticipating recurring errors or context switches before they occur. Although \serum's computational complexity presents a challenge in latency to its feasibility in live settings, the majority of compute will be front loaded into a startup cost as \serum learns a user's workflow, with minor revisions after the incubation period. This frees up compute for live suggestions. 
Evaluation in live assistive settings and scaling to larger video corpora are the
highest-priority directions for future work.

\textbf{Counterfactual scenarios.}
Future work could explore reversing \serum to allow video generation models to imagine counterfactual scenarios. While \serum infers actions and intentions from video, the reverse would use \serum's action and intent labels to generate video. This would enable generating counterfactual videos through perturbing inferred actions and intentions. 

\textbf{Hallucinations.}
Over \serum's iterative annotation passes, we observe two main sources of hallucinations: 1. The image is not clear (e.g., due to motion blur, occlusion). 2. The VLM confuses whether an action is starting or ending due to limited temporal granularity. For most hallucinations \serum self-corrects by re-examining the original frame in each pass and by attaining neighbor consensus via the temporal context window to normalize inconsistent cases.

\section*{Acknowledgements.}
We thank the members of the Minnesota NLP group for giving feedback on initial drafts and, crucially, our colleague-annotators (Khanh Chi Le, Ruizi Wang, Jingcheng Liang) who dedicated significant time annotating \serum's results over several trials.

\section*{Ethics Statement}
This work analyzes publicly available YouTube videos and does not involve human subjects research. We acknowledge that behavioral modeling from screen recordings could be misused for unauthorized surveillance; our work is intended for user-initiated workflow analysis and support. We release our code to promote reproducibility and encourage its responsible use. Annotations are generated by a vision-language model and may reflect biases present in its training data. \footnote{Additional LLM Disclosure in Appendix ~\ref{appendix:appendix_llm_disclosure} }

\textbf{Human annotation.} Two forms of human annotation supported this work:
(1) a calibration set of 200 label pairs (100 activities, 100 intents) was 
hand-rated by one of the authors to fit the SentenceBERT semantic-merge threshold $t^*$
(\S\ref{tab:canon_summary}); (2) five members of our research lab rated
final-pass labels for accuracy on a set of 26 videos pre-vetted by 
the authors (\S\ref{sec:rq3_human}). Annotators were uncompensated lab 
volunteers, viewed only the pre-vetted videos, and agreed to participate and to the use of their judgements in this research. No personally identifying information was
collected from annotators, and the videos contained no third-party private 
data. We did not seek formal Institutional Review Board approval, treating 
the rating task as internal validation by research collaborators; we 
acknowledge this is a limitation of the human evaluation and that a small, 
in-lab annotator pool may bias results toward positive judgments.

\bibliography{colm2026_conference}
\bibliographystyle{colm2026_conference}

\appendix
\section{Full Data Collection}
\label{appendix:full_data_collection}
\begin{longtable}{llrrp{3.6cm}}
\caption{Dataset Overview}\label{tab:dataset-overview}\\
\toprule
Video & Category & Frames & Passes & Source URL \\
\midrule
\endfirsthead
\caption[]{Dataset Overview (continued)}\\
\toprule
Video & Category & Frames & Passes & Source URL \\
\midrule
\endhead
\endfoot
\bottomrule
\endlastfoot
ACS\_salestrainingP12 & Daily Life & 55 & 12 & \url{youtu.be/ZG4ExqMVA7w} \\
AC\_leetcode2P12 & Coding & 80 & 12 & \url{youtu.be/vRAK2YnFr1o} \\
AC\_leetcodeP12 & Coding & 352 & 12 & \url{youtu.be/zeLZuhi6eYU} \\
AC\_pizzaP12 & Cooking & 86 & 12 & \url{youtu.be/Q9j6HhF0tGE} \\
AC\_profreactsP12 & Daily Life & 108 & 12 & \url{youtu.be/3mRvCF4qyTA} \\
AC\_sandwichP12 & Cooking & 120 & 12 & \url{youtu.be/ad8TWumCSnY} \\
AC\_studrecordingP12 & Daily Life & 100 & 12 & \url{youtu.be/eB54LIupAhU} \\
AC\_ukdayinlifeP12 & Daily Life & 187 & 12 & \url{youtu.be/BvWnEiOoAEk} \\
AC\_waiterP12 & Cooking & 133 & 12 & \url{youtu.be/w4pGt-iGpBI} \\
BC\_dunkinhelpP12 & Cooking & 213 & 12 & \url{youtu.be/j_gUBLwxG1U} \\
BC\_nycswevlogP12 & Daily Life & 114 & 12 & \url{youtu.be/4lo81zt7HK8} \\
BC\_pizzarushP12 & Cooking & 252 & 12 & \url{youtu.be/S5ltPbUur38} \\
BC\_swevlogP12 & Daily Life & 121 & 12 & \url{youtu.be/b_eeMSNO97U} \\
BC\_vibecodingP12 & Coding & 157 & 12 & \url{youtu.be/P3JA7MTiGg8} \\
CC\_baristaP12 & Cooking & 429 & 12 & \url{youtu.be/jdguVU0F7fs} \\
CC\_swisssweP12 & Daily Life & 327 & 12 & \url{youtu.be/_GSI2RaiV0s} \\
DC\_calcappcodingP12 & Coding & 409 & 12 & \url{youtu.be/sBJmRD7kNTk} \\
DC\_snakecodingP12 & Coding & 289 & 12 & \url{youtu.be/Wlu4MsBnjuk} \\
PERS\_coinflipP12 & Coding & 137 & 12 & \url{youtu.be/-o-H1Ecqo_M} \\
PERS\_movieP12 & Coding & 200 & 12 & \url{youtu.be/J6uam9jEmDU} \\
PERS\_weatherP12 & Coding & 246 & 12 & \url{youtu.be/iILFBGm_I9M} \\
bartenderP12 & Cooking & 212 & 12 & \url{youtu.be/1G-9Pibx5JI} \\
basketballP12 & Physical & 50 & 12 & \url{youtu.be/N7RNoleA7Sk} \\
behindP12 & Daily Life & 124 & 12 & \url{youtu.be/h4exLX8Wz4E} \\
carrepair2P12 & Physical & 113 & 12 & \url{youtu.be/o0OBJCfAfOY} \\
carrepair3P12 & Physical & 171 & 12 & \url{youtu.be/VdR5zPyqp_4} \\
carrepairP12 & Physical & 223 & 12 & \url{youtu.be/vHdz74orr1Q} \\
cashboothP12 & Cooking & 50 & 12 & \url{youtu.be/9lNBUsF4WRU} \\
coding2P12 & Coding & 185 & 12 & \url{youtu.be/gRyvG7PZ4m0} \\
coding3P12 & Coding & 226 & 12 & \url{youtu.be/825u2Puaej0} \\
codingP12 & Coding & 179 & 12 & \url{youtu.be/DfDPJqD3FjI} \\
codinglogoP12 & Coding & 234 & 12 & \url{youtu.be/B_puD1rTsOQ} \\
codingqrcodeP12 & Coding & 198 & 12 & \url{youtu.be/I50Xwve6QW4} \\
competitiveP12 & Coding & 206 & 12 & \url{youtu.be/uGrBHohIgQY} \\
compgamingP12 & Coding & 127 & 12 & \url{youtu.be/yCezqhatLV8} \\
construction2P12 & Physical & 81 & 12 & \url{youtu.be/GlsCRChrdfU} \\
constructionP12 & Physical & 260 & 12 & \url{youtu.be/2avgoVsQ_og} \\
csscodingP12 & Coding & 42 & 12 & \url{youtu.be/EZhPsuIXawk} \\
dayinthelifesweP12 & Daily Life & 102 & 12 & \url{youtu.be/aTHBJwVgu3I} \\
drivingP12 & Physical & 158 & 12 & \url{youtu.be/iSnP5c997Uk} \\
dunkinP12 & Cooking & 210 & 12 & \url{youtu.be/hEJaSuDiQU8} \\
fluttercodingP12 & Coding & 216 & 12 & \url{youtu.be/C7Kafde7gZ4} \\
goprochefP12 & Cooking & 299 & 12 & \url{youtu.be/CBSsL4u_nng} \\
headchefP12 & Cooking & 350 & 12 & \url{youtu.be/Ipe9xJCfuTM} \\
hotdogP12 & Cooking & 265 & 12 & \url{youtu.be/YMpGWAB41lI} \\
labworkP12 & Daily Life & 164 & 12 & \url{youtu.be/C3aKnhXn20U} \\
markiplierP12 & Daily Life & 317 & 12 & \url{youtu.be/Yk-I7IVLAGo} \\
mcdcookP12 & Cooking & 151 & 12 & \url{youtu.be/8kcUsQdxtSs} \\
mcdtakingordersP12 & Cooking & 337 & 12 & \url{youtu.be/_c8PppBiMqE} \\
microbialP12 & Daily Life & 54 & 12 & \url{youtu.be/NUkrCXMdl3o} \\
musicplayercodingP12 & Coding & 283 & 12 & \url{youtu.be/KndQpfPkOOY} \\
paperworkP12 & Daily Life & 69 & 12 & \url{youtu.be/JdkMmLhPw_E} \\
phonerepairP12 & Physical & 343 & 12 & \url{youtu.be/p9hA59nn7uQ} \\
radiatorrepairP12 & Physical & 98 & 12 & \url{youtu.be/ldIo1L6S_Sw} \\
rmlineP12 & Physical & 105 & 12 & \url{youtu.be/jIJTEm0qNuo} \\
sushiP12 & Cooking & 183 & 12 & \url{youtu.be/KUzYFMgWs4w} \\
tractorfarmingP12 & Physical & 96 & 12 & \url{youtu.be/rqA-iT2DKO4} \\
tttcodingP12 & Coding & 175 & 12 & \url{youtu.be/MgtGHfdpigU} \\
tutorialP12 & Daily Life & 92 & 12 & \url{youtu.be/a32fbqPNir4} \\
welshgardeningP12 & Physical & 160 & 12 & \url{youtu.be/T3fgL091hXs} \\
wslinstallP12 & Daily Life & 102 & 12 & \url{youtu.be/QadguqFAt_8} \\
\end{longtable}

\clearpage
\section{Verbose Next-Action Prediction Task Results}
\label{appendix:verbose_NA}

\footnotesize
\setlength{\tabcolsep}{2pt}
\begin{longtable}{ll|rrrrr|rrr}
\caption{Markov prediction accuracy, final pass. $V$ = vocabulary size; $^n$ = normalized labels.} \label{tab:markov-final} \\
\toprule
Video & Type & V & Markov & Majority & Wt.~Rand & Uniform & V$^n$ & Markov$^n$ & Maj$^n$ \\
\midrule
\endfirsthead

\multicolumn{10}{c}{\small\tablename~\thetable{} -- continued} \\
\toprule
Video & Type & V & Markov & Majority & Wt.~Rand & Uniform & V$^n$ & Markov$^n$ & Maj$^n$ \\
\midrule
\endhead

ACS\_salestrainingP12 & Act & 7 & 61.9\% & 19.0\% & 28.6\% & 14.3\% & 6 & 61.9\% & 19.0\% \\
ACS\_salestrainingP12 & Int & 24 & 4.8\% & 0.0\% & 4.3\% & 4.2\% & 7 & 9.5\% & 38.1\% \\
AC\_leetcode2P12 & Act & 2 & 100.0\% & 100.0\% & 96.0\% & 50.0\% & 2 & 100.0\% & 100.0\% \\
AC\_leetcode2P12 & Int & 8 & 83.9\% & 83.9\% & 40.0\% & 12.5\% & 5 & 100.0\% & 100.0\% \\
AC\_leetcodeP12 & Act & 6 & 92.6\% & 94.3\% & 84.5\% & 16.7\% & 5 & 93.4\% & 94.3\% \\
AC\_leetcodeP12 & Int & 11 & 63.1\% & 69.7\% & 41.0\% & 9.1\% & 6 & 65.6\% & 69.7\% \\
AC\_pizzaP12 & Act & 45 & 5.9\% & 0.0\% & 2.1\% & 2.2\% & 25 & 8.8\% & 0.0\% \\
AC\_pizzaP12 & Int & 43 & 0.0\% & 5.9\% & 2.8\% & 2.3\% & 13 & 52.9\% & 32.4\% \\
AC\_profreactsP12 & Act & 22 & 11.6\% & 20.9\% & 10.5\% & 4.5\% & 19 & 16.3\% & 20.9\% \\
AC\_profreactsP12 & Int & 40 & 7.0\% & 23.3\% & 4.4\% & 2.5\% & 26 & 14.0\% & 25.6\% \\
AC\_sandwichP12 & Act & 46 & 4.3\% & 0.0\% & 1.4\% & 2.2\% & 23 & 12.8\% & 12.8\% \\
AC\_sandwichP12 & Int & 19 & 51.1\% & 48.9\% & 30.3\% & 5.3\% & 7 & 74.5\% & 74.5\% \\
AC\_studrecordingP12 & Act & 20 & 48.7\% & 25.6\% & 8.6\% & 5.0\% & 14 & 51.3\% & 7.7\% \\
AC\_studrecordingP12 & Int & 19 & 59.0\% & 23.1\% & 10.4\% & 5.3\% & 8 & 64.1\% & 28.2\% \\
AC\_ukdayinlifeP12 & Act & 44 & 12.2\% & 5.4\% & 3.3\% & 2.3\% & 31 & 24.3\% & 5.4\% \\
AC\_ukdayinlifeP12 & Int & 66 & 5.4\% & 1.4\% & 1.9\% & 1.5\% & 37 & 17.6\% & 9.5\% \\
AC\_waiterP12 & Act & 24 & 20.8\% & 24.5\% & 10.2\% & 4.2\% & 15 & 32.1\% & 39.6\% \\
AC\_waiterP12 & Int & 18 & 28.3\% & 39.6\% & 20.4\% & 5.6\% & 7 & 56.6\% & 47.2\% \\
BC\_dunkinhelpP12 & Act & 66 & 17.6\% & 25.9\% & 6.4\% & 1.5\% & 29 & 50.6\% & 52.9\% \\
BC\_dunkinhelpP12 & Int & 42 & 15.3\% & 22.4\% & 9.5\% & 2.4\% & 13 & 65.9\% & 71.8\% \\
BC\_nycswevlogP12 & Act & 50 & 6.7\% & 6.7\% & 2.8\% & 2.0\% & 45 & 6.7\% & 6.7\% \\
BC\_nycswevlogP12 & Int & 59 & 6.7\% & 4.4\% & 1.6\% & 1.7\% & 47 & 6.7\% & 4.4\% \\
BC\_pizzarushP12 & Act & 67 & 31.0\% & 5.0\% & 5.0\% & 1.5\% & 28 & 52.0\% & 13.0\% \\
BC\_pizzarushP12 & Int & 38 & 53.0\% & 26.0\% & 15.1\% & 2.6\% & 15 & 86.0\% & 87.0\% \\
BC\_swevlogP12 & Act & 31 & 12.5\% & 20.8\% & 7.3\% & 3.2\% & 24 & 29.2\% & 22.9\% \\
BC\_swevlogP12 & Int & 43 & 6.2\% & 8.3\% & 4.1\% & 2.3\% & 21 & 8.3\% & 8.3\% \\
BC\_vibecodingP12 & Act & 9 & 43.5\% & 50.0\% & 31.5\% & 11.1\% & 8 & 62.9\% & 50.0\% \\
BC\_vibecodingP12 & Int & 18 & 40.3\% & 43.5\% & 31.2\% & 5.6\% & 8 & 48.4\% & 43.5\% \\
CC\_baristaP12 & Act & 94 & 18.1\% & 21.1\% & 6.2\% & 1.1\% & 37 & 32.7\% & 28.7\% \\
CC\_baristaP12 & Int & 83 & 7.6\% & 0.0\% & 2.5\% & 1.2\% & 20 & 48.5\% & 33.3\% \\
CC\_swisssweP12 & Act & 89 & 5.4\% & 7.7\% & 2.2\% & 1.1\% & 53 & 8.5\% & 8.5\% \\
CC\_swisssweP12 & Int & 105 & 20.0\% & 21.5\% & 2.6\% & 1.0\% & 64 & 24.6\% & 21.5\% \\
DC\_calcappcodingP12 & Act & 6 & 82.2\% & 82.8\% & 55.2\% & 16.7\% & 6 & 82.2\% & 82.8\% \\
DC\_calcappcodingP12 & Int & 86 & 1.8\% & 0.0\% & 0.5\% & 1.2\% & 19 & 90.8\% & 91.4\% \\
DC\_snakecodingP12 & Act & 5 & 10.4\% & 11.3\% & 11.5\% & 20.0\% & 3 & 11.3\% & 11.3\% \\
DC\_snakecodingP12 & Int & 9 & 11.3\% & 14.8\% & 13.0\% & 11.1\% & 2 & 99.1\% & 99.1\% \\
PERS\_coinflipP12 & Act & 10 & 63.0\% & 59.3\% & 45.0\% & 10.0\% & 6 & 72.2\% & 59.3\% \\
PERS\_coinflipP12 & Int & 33 & 57.4\% & 0.0\% & 4.6\% & 3.0\% & 10 & 92.6\% & 94.4\% \\
PERS\_movieP12 & Act & 6 & 100.0\% & 100.0\% & 88.9\% & 16.7\% & 5 & 100.0\% & 100.0\% \\
PERS\_movieP12 & Int & 36 & 43.0\% & 8.9\% & 8.5\% & 2.8\% & 9 & 72.2\% & 13.9\% \\
PERS\_weatherP12 & Act & 8 & 63.3\% & 62.2\% & 33.1\% & 12.5\% & 5 & 63.3\% & 62.2\% \\
PERS\_weatherP12 & Int & 28 & 34.7\% & 0.0\% & 8.4\% & 3.6\% & 4 & 95.9\% & 95.9\% \\
bartenderP12 & Act & 85 & 10.7\% & 2.4\% & 1.9\% & 1.2\% & 45 & 11.9\% & 3.6\% \\
bartenderP12 & Int & 96 & 10.7\% & 11.9\% & 1.8\% & 1.0\% & 40 & 28.6\% & 25.0\% \\
basketballP12 & Act & 10 & 21.1\% & 21.1\% & 13.8\% & 10.0\% & 7 & 42.1\% & 52.6\% \\
basketballP12 & Int & 9 & 89.5\% & 89.5\% & 41.8\% & 11.1\% & 5 & 100.0\% & 100.0\% \\
behindP12 & Act & 15 & 12.2\% & 0.0\% & 3.9\% & 6.7\% & 14 & 12.2\% & 0.0\% \\
behindP12 & Int & 19 & 8.2\% & 0.0\% & 2.2\% & 5.3\% & 13 & 8.2\% & 0.0\% \\
carrepair2P12 & Act & 32 & 8.9\% & 13.3\% & 4.6\% & 3.1\% & 19 & 8.9\% & 13.3\% \\
carrepair2P12 & Int & 31 & 31.1\% & 44.4\% & 12.8\% & 3.2\% & 14 & 53.3\% & 57.8\% \\
carrepair3P12 & Act & 44 & 2.9\% & 5.9\% & 2.2\% & 2.3\% & 20 & 20.6\% & 22.1\% \\
carrepair3P12 & Int & 38 & 13.2\% & 11.8\% & 8.5\% & 2.6\% & 16 & 29.4\% & 38.2\% \\
carrepairP12 & Act & 52 & 1.1\% & 0.0\% & 1.4\% & 1.9\% & 34 & 2.2\% & 0.0\% \\
carrepairP12 & Int & 33 & 32.6\% & 30.3\% & 12.7\% & 3.0\% & 15 & 46.1\% & 47.2\% \\
cashboothP12 & Act & 10 & 10.5\% & 21.1\% & 10.5\% & 10.0\% & 8 & 15.8\% & 26.3\% \\
cashboothP12 & Int & 14 & 31.6\% & 10.5\% & 12.0\% & 7.1\% & 6 & 42.1\% & 47.4\% \\
coding2P12 & Act & 14 & 8.2\% & 0.0\% & 6.2\% & 7.1\% & 9 & 16.4\% & 0.0\% \\
coding2P12 & Int & 62 & 8.2\% & 0.0\% & 2.7\% & 1.6\% & 16 & 16.4\% & 0.0\% \\
coding3P12 & Act & 10 & 68.9\% & 66.7\% & 54.3\% & 10.0\% & 7 & 66.7\% & 66.7\% \\
coding3P12 & Int & 8 & 90.0\% & 90.0\% & 77.1\% & 12.5\% & 5 & 100.0\% & 100.0\% \\
codingP12 & Act & 4 & 100.0\% & 100.0\% & 91.0\% & 25.0\% & 3 & 100.0\% & 100.0\% \\
codingP12 & Int & 17 & 42.3\% & 16.9\% & 16.5\% & 5.9\% & 4 & 90.1\% & 90.1\% \\
codinglogoP12 & Act & 2 & 98.9\% & 98.9\% & 98.2\% & 50.0\% & 2 & 98.9\% & 98.9\% \\
codinglogoP12 & Int & 2 & 100.0\% & 100.0\% & 98.6\% & 50.0\% & 2 & 100.0\% & 100.0\% \\
codingqrcodeP12 & Act & 8 & 89.9\% & 93.7\% & 76.6\% & 12.5\% & 6 & 89.9\% & 93.7\% \\
codingqrcodeP12 & Int & 42 & 32.9\% & 48.1\% & 10.5\% & 2.4\% & 9 & 60.8\% & 75.9\% \\
competitiveP12 & Act & 5 & 72.0\% & 76.8\% & 72.8\% & 20.0\% & 4 & 73.2\% & 76.8\% \\
competitiveP12 & Int & 12 & 37.8\% & 26.8\% & 21.3\% & 8.3\% & 3 & 68.3\% & 74.4\% \\
compgamingP12 & Act & 3 & 96.0\% & 96.0\% & 84.1\% & 33.3\% & 3 & 96.0\% & 96.0\% \\
compgamingP12 & Int & 2 & 76.0\% & 76.0\% & 62.0\% & 50.0\% & 1 & -- & -- \\
construction2P12 & Act & 24 & 21.9\% & 28.1\% & 10.5\% & 4.2\% & 12 & 21.9\% & 31.2\% \\
construction2P12 & Int & 26 & 12.5\% & 34.4\% & 9.2\% & 3.8\% & 5 & 71.9\% & 71.9\% \\
constructionP12 & Act & 13 & 71.8\% & 75.7\% & 49.9\% & 7.7\% & 9 & 72.8\% & 75.7\% \\
constructionP12 & Int & 18 & 22.0\% & 26.8\% & 21.9\% & 5.6\% & 7 & 81.7\% & 81.7\% \\
csscodingP12 & Act & 3 & 87.5\% & 87.5\% & 50.4\% & 33.3\% & 2 & 93.8\% & 87.5\% \\
csscodingP12 & Int & 8 & 37.5\% & 0.0\% & 19.1\% & 12.5\% & 4 & 62.5\% & 62.5\% \\
dayinthelifesweP12 & Act & 42 & 0.0\% & 7.5\% & 2.2\% & 2.4\% & 31 & 5.0\% & 12.5\% \\
dayinthelifesweP12 & Int & 48 & 12.5\% & 15.0\% & 2.3\% & 2.1\% & 31 & 12.5\% & 0.0\% \\
drivingP12 & Act & 21 & 81.0\% & 84.1\% & 25.5\% & 4.8\% & 14 & 85.7\% & 92.1\% \\
drivingP12 & Int & 17 & 93.7\% & 95.2\% & 46.7\% & 5.9\% & 11 & 95.2\% & 95.2\% \\
dunkinP12 & Act & 98 & 1.2\% & 2.4\% & 1.3\% & 1.0\% & 39 & 10.8\% & 13.3\% \\
dunkinP12 & Int & 74 & 6.0\% & 14.5\% & 2.9\% & 1.4\% & 25 & 10.8\% & 22.9\% \\
fluttercodingP12 & Act & 14 & 62.8\% & 68.6\% & 43.8\% & 7.1\% & 9 & 65.1\% & 68.6\% \\
fluttercodingP12 & Int & 55 & 30.2\% & 37.2\% & 5.9\% & 1.8\% & 22 & 45.3\% & 54.7\% \\
goprochefP12 & Act & 112 & 3.4\% & 10.1\% & 2.3\% & 0.9\% & 53 & 21.8\% & 31.9\% \\
goprochefP12 & Int & 68 & 14.3\% & 21.8\% & 6.5\% & 1.5\% & 25 & 23.5\% & 27.7\% \\
headchefP12 & Act & 81 & 15.1\% & 18.0\% & 4.4\% & 1.2\% & 35 & 23.0\% & 19.4\% \\
headchefP12 & Int & 59 & 34.5\% & 42.4\% & 15.1\% & 1.7\% & 22 & 39.6\% & 42.4\% \\
hotdogP12 & Act & 70 & 18.1\% & 24.8\% & 5.4\% & 1.4\% & 30 & 54.3\% & 48.6\% \\
hotdogP12 & Int & 48 & 12.4\% & 19.0\% & 6.9\% & 2.1\% & 15 & 82.9\% & 83.8\% \\
labworkP12 & Act & 8 & 67.7\% & 67.7\% & 44.8\% & 12.5\% & 6 & 70.8\% & 70.8\% \\
labworkP12 & Int & 9 & 46.2\% & 38.5\% & 27.1\% & 11.1\% & 2 & 47.7\% & 47.7\% \\
markiplierP12 & Act & 22 & 46.0\% & 49.2\% & 30.1\% & 4.5\% & 15 & 66.7\% & 67.5\% \\
markiplierP12 & Int & 53 & 19.0\% & 23.8\% & 9.2\% & 1.9\% & 26 & 39.7\% & 46.0\% \\
mcdcookP12 & Act & 30 & 25.0\% & 26.7\% & 8.9\% & 3.3\% & 9 & 80.0\% & 80.0\% \\
mcdcookP12 & Int & 21 & 28.3\% & 38.3\% & 14.1\% & 4.8\% & 4 & 95.0\% & 95.0\% \\
mcdtakingordersP12 & Act & 63 & 20.9\% & 16.4\% & 5.1\% & 1.6\% & 42 & 29.9\% & 24.6\% \\
mcdtakingordersP12 & Int & 66 & 11.9\% & 1.5\% & 3.5\% & 1.5\% & 35 & 21.6\% & 4.5\% \\
microbialP12 & Act & 17 & 4.8\% & 0.0\% & 8.1\% & 5.9\% & 9 & 19.0\% & 19.0\% \\
microbialP12 & Int & 6 & 52.4\% & 61.9\% & 45.6\% & 16.7\% & 2 & 90.5\% & 95.2\% \\
musicplayercodingP12 & Act & 8 & 92.9\% & 94.7\% & 83.5\% & 12.5\% & 5 & 92.9\% & 94.7\% \\
musicplayercodingP12 & Int & 43 & 11.5\% & 10.6\% & 4.3\% & 2.3\% & 9 & 61.9\% & 12.4\% \\
paperworkP12 & Act & 15 & 11.1\% & 11.1\% & 8.8\% & 6.7\% & 8 & 33.3\% & 29.6\% \\
paperworkP12 & Int & 7 & 22.2\% & 33.3\% & 27.8\% & 14.3\% & 3 & 44.4\% & 44.4\% \\
phonerepairP12 & Act & 61 & 15.3\% & 17.5\% & 5.8\% & 1.6\% & 30 & 20.4\% & 19.0\% \\
phonerepairP12 & Int & 31 & 34.3\% & 10.9\% & 9.6\% & 3.2\% & 15 & 56.9\% & 64.2\% \\
radiatorrepairP12 & Act & 9 & 53.8\% & 64.1\% & 30.7\% & 11.1\% & 5 & 76.9\% & 84.6\% \\
radiatorrepairP12 & Int & 5 & 41.0\% & 46.2\% & 36.8\% & 20.0\% & 3 & 100.0\% & 100.0\% \\
rmlineP12 & Act & 10 & 7.3\% & 2.4\% & 11.6\% & 10.0\% & 5 & 48.8\% & 51.2\% \\
rmlineP12 & Int & 10 & 65.9\% & 68.3\% & 34.3\% & 10.0\% & 3 & 82.9\% & 82.9\% \\
sushiP12 & Act & 92 & 1.4\% & 6.8\% & 1.2\% & 1.1\% & 26 & 39.7\% & 50.7\% \\
sushiP12 & Int & 38 & 21.9\% & 20.5\% & 8.3\% & 2.6\% & 11 & 78.1\% & 79.5\% \\
tractorfarmingP12 & Act & 25 & 7.9\% & 0.0\% & 4.0\% & 4.0\% & 18 & 78.9\% & 84.2\% \\
tractorfarmingP12 & Int & 34 & 2.6\% & 0.0\% & 2.9\% & 2.9\% & 19 & 81.6\% & 0.0\% \\
tttcodingP12 & Act & 6 & 72.5\% & 72.5\% & 54.9\% & 16.7\% & 6 & 72.5\% & 72.5\% \\
tttcodingP12 & Int & 13 & 100.0\% & 100.0\% & 75.4\% & 7.7\% & 4 & 100.0\% & 100.0\% \\
tutorialP12 & Act & 13 & 47.2\% & 52.8\% & 15.8\% & 7.7\% & 10 & 72.2\% & 58.3\% \\
tutorialP12 & Int & 31 & 50.0\% & 41.7\% & 8.6\% & 3.2\% & 16 & 63.9\% & 16.7\% \\
welshgardeningP12 & Act & 42 & 25.4\% & 38.1\% & 7.7\% & 2.4\% & 20 & 60.3\% & 68.3\% \\
welshgardeningP12 & Int & 39 & 44.4\% & 19.0\% & 8.7\% & 2.6\% & 14 & 58.7\% & 28.6\% \\
wslinstallP12 & Act & 27 & 22.5\% & 42.5\% & 10.3\% & 3.7\% & 21 & 25.0\% & 47.5\% \\
wslinstallP12 & Int & 49 & 2.5\% & 5.0\% & 1.5\% & 2.0\% & 24 & 2.5\% & 0.0\% \\
\end{longtable}

\clearpage
\section{Generalizing Procedure on EPIC-KITCHENS-100}
\label{appendix:appendix_generalization}


\begin{table}[h]
\centering
\small
\caption{EPIC-KITCHENS-100 generalization (366 videos from 37 participants). Same Markov harness as Table~\ref{tab:cross_video_accuracy}; final-pass P11 (activity) / P12 (intent). $^{n}$ denotes models built on normalized labels.}
\label{tab:cross_video_accuracy_ek}
\setlength{\tabcolsep}{5pt}
\begin{tabular}{lrrrr}
\toprule
& \multicolumn{2}{c}{Activity} & \multicolumn{2}{c}{Intent} \\
\cmidrule(lr){2-3}\cmidrule(lr){4-5}
Model & Top-1~($\uparrow$) & PPL~($\downarrow$) & Top-1~($\uparrow$) & PPL~($\downarrow$) \\
\midrule
Markov & 15.3{\small$\pm$18.9} & 27.4{\small$\pm$13.5} & 33.1{\small$\pm$27.6} & 14.4{\small$\pm$9.1} \\
Majority & 14.5{\small$\pm$20.1} & 42.2{\small$\pm$25.3} & 31.6{\small$\pm$29.5} & 20.4{\small$\pm$16.6} \\
Wt.\ Random & 6.4{\small$\pm$7.6} & 42.2{\small$\pm$25.3} & 17.6{\small$\pm$17.7} & 20.4{\small$\pm$16.6} \\
Uniform & 4.1{\small$\pm$2.6} & 31.2{\small$\pm$12.7} & 7.4{\small$\pm$6.3} & 19.1{\small$\pm$8.9} \\
\midrule
Markov$^{n}$ & 28.3{\small$\pm$25.0} & 13.8{\small$\pm$7.6} & 53.6{\small$\pm$28.7} & 5.5{\small$\pm$3.5} \\
Majority$^{n}$ & 22.4{\small$\pm$25.4} & 21.5{\small$\pm$15.1} & 47.0{\small$\pm$32.9} & 7.7{\small$\pm$6.9} \\
\bottomrule
\end{tabular}
\\[2pt]\footnotesize 366 videos, 33{,}788 frames at the final activity pass.
\end{table}

\begin{table}[h]
\centering
\small
\caption{Curated vs.\ EPIC-KITCHENS-100 generalization (61 curated videos vs.\ 366 EK videos; same Markov harness). Absolute Markov$^{n}$ accuracy is lower on EK; the Markov$^{n}$$-$Majority$^{n}$ method gap is wider on EK. $\Delta = \text{EK} - \text{Curated}$. $^{n}$ denotes models built on normalized labels.}
\label{tab:curated_vs_ek}
\setlength{\tabcolsep}{6pt}
\begin{tabular}{lrrr}
\toprule
Metric & Curated & EK & $\Delta$ \\
\midrule
\multicolumn{4}{l}{\textit{Absolute Markov$^{n}$ performance (curated wins on accuracy; EK wins on intent PPL):}} \\
\quad Activity Top-1 (\%) & 47.0{\small$\pm$28.5} & 28.3{\small$\pm$25.0} & -18.7\,pp \\
\quad Activity PPL & 10.4{\small$\pm$9.5} & 13.8{\small$\pm$7.6} & +3.5 \\
\quad Intent Top-1 (\%) & 58.1{\small$\pm$30.7} & 53.6{\small$\pm$28.7} & -4.5\,pp \\
\quad Intent PPL & 7.6{\small$\pm$9.1} & 5.5{\small$\pm$3.5} & -2.1 \\
\midrule
\multicolumn{4}{l}{\textit{Markov$^{n}$$-$Majority$^{n}$ method gap (EK gap is wider on both):}} \\
\quad Activity (pp) & +3.3 & +5.9 & +2.6\,pp \\
\quad Intent (pp) & +3.4 & +6.5 & +3.2\,pp \\
\bottomrule
\end{tabular}
\end{table}

\begin{longtable}{lrrrrr}
\caption{Per-participant Markov$^{n}$ top-1 accuracy and perplexity on EPIC-KITCHENS-100 (final-pass P11/P12, normalized labels).}
\label{tab:ek_per_participant} \\
\toprule
Participant & Videos & Act\,Top-1 & Act\,PPL & Int\,Top-1 & Int\,PPL \\
\midrule
\endfirsthead
\toprule
Participant & Videos & Act\,Top-1 & Act\,PPL & Int\,Top-1 & Int\,PPL \\
\midrule
\endhead
P04 & 28 & 20.9{\small$\pm$23.5} & 15.4{\small$\pm$8.1} & 42.3{\small$\pm$26.9} & 7.2{\small$\pm$3.7} \\
P22 & 27 & 23.3{\small$\pm$14.8} & 16.4{\small$\pm$7.1} & 45.3{\small$\pm$23.9} & 6.6{\small$\pm$4.0} \\
P02 & 23 & 28.6{\small$\pm$30.8} & 15.7{\small$\pm$7.8} & 59.3{\small$\pm$30.2} & 5.3{\small$\pm$3.5} \\
P03 & 23 & 27.6{\small$\pm$22.2} & 11.2{\small$\pm$5.2} & 58.2{\small$\pm$26.8} & 4.7{\small$\pm$3.0} \\
P08 & 17 & 14.8{\small$\pm$11.6} & 18.6{\small$\pm$8.7} & 51.0{\small$\pm$28.2} & 6.5{\small$\pm$3.5} \\
P28 & 17 & 32.8{\small$\pm$18.3} & 11.9{\small$\pm$6.6} & 53.1{\small$\pm$34.8} & 5.1{\small$\pm$3.6} \\
P30 & 17 & 15.8{\small$\pm$16.5} & 18.2{\small$\pm$8.0} & 36.5{\small$\pm$29.6} & 7.6{\small$\pm$3.9} \\
P01 & 16 & 29.9{\small$\pm$26.7} & 14.5{\small$\pm$7.3} & 58.9{\small$\pm$27.3} & 5.0{\small$\pm$3.0} \\
P07 & 16 & 22.3{\small$\pm$15.9} & 12.0{\small$\pm$7.0} & 56.6{\small$\pm$30.8} & 4.8{\small$\pm$3.6} \\
P26 & 16 & 42.2{\small$\pm$27.3} & 7.0{\small$\pm$3.8} & 70.9{\small$\pm$28.6} & 2.7{\small$\pm$1.7} \\
P06 & 13 & 34.0{\small$\pm$29.7} & 13.2{\small$\pm$6.5} & 48.5{\small$\pm$30.9} & 5.7{\small$\pm$3.2} \\
P25 & 12 & 28.2{\small$\pm$22.7} & 12.0{\small$\pm$5.6} & 51.2{\small$\pm$22.2} & 5.4{\small$\pm$2.8} \\
P11 & 10 & 19.8{\small$\pm$15.8} & 15.8{\small$\pm$4.1} & 47.0{\small$\pm$24.8} & 6.2{\small$\pm$3.1} \\
P12 & 10 & 29.7{\small$\pm$12.8} & 13.3{\small$\pm$4.7} & 38.7{\small$\pm$24.9} & 6.8{\small$\pm$3.5} \\
P27 & 10 & 32.0{\small$\pm$27.6} & 13.5{\small$\pm$9.5} & 51.1{\small$\pm$29.1} & 6.2{\small$\pm$4.4} \\
P31 & 9 & 23.4{\small$\pm$29.7} & 17.2{\small$\pm$9.4} & 64.0{\small$\pm$24.4} & 4.6{\small$\pm$2.2} \\
P33 & 9 & 25.4{\small$\pm$22.8} & 14.9{\small$\pm$5.4} & 41.3{\small$\pm$26.7} & 6.4{\small$\pm$2.3} \\
P35 & 9 & 33.6{\small$\pm$23.6} & 14.0{\small$\pm$8.4} & 56.3{\small$\pm$21.5} & 4.7{\small$\pm$2.2} \\
P15 & 8 & 24.1{\small$\pm$14.5} & 15.1{\small$\pm$6.9} & 56.8{\small$\pm$25.7} & 4.9{\small$\pm$2.9} \\
P09 & 7 & 28.3{\small$\pm$26.1} & 12.0{\small$\pm$6.7} & 46.2{\small$\pm$29.5} & 5.7{\small$\pm$3.4} \\
P23 & 7 & 43.7{\small$\pm$37.8} & 10.7{\small$\pm$7.3} & 56.2{\small$\pm$30.7} & 5.4{\small$\pm$3.6} \\
P24 & 7 & 36.4{\small$\pm$17.1} & 12.0{\small$\pm$6.2} & 62.3{\small$\pm$26.9} & 3.9{\small$\pm$2.1} \\
P34 & 7 & 42.8{\small$\pm$26.2} & 8.8{\small$\pm$4.4} & 61.3{\small$\pm$26.7} & 4.2{\small$\pm$2.4} \\
P05 & 6 & 47.4{\small$\pm$29.0} & 9.4{\small$\pm$5.5} & 79.0{\small$\pm$16.0} & 2.8{\small$\pm$1.3} \\
P18 & 6 & 26.1{\small$\pm$26.8} & 14.7{\small$\pm$7.7} & 47.0{\small$\pm$15.7} & 6.5{\small$\pm$3.7} \\
P20 & 5 & 18.2{\small$\pm$17.4} & 17.8{\small$\pm$6.4} & 65.4{\small$\pm$23.5} & 4.1{\small$\pm$1.8} \\
P13 & 4 & 64.7{\small$\pm$29.0} & 7.8{\small$\pm$4.7} & 66.9{\small$\pm$32.8} & 4.8{\small$\pm$4.4} \\
P29 & 4 & 26.1{\small$\pm$34.1} & 13.2{\small$\pm$8.8} & 71.8{\small$\pm$26.5} & 3.6{\small$\pm$3.0} \\
P32 & 4 & 31.1{\small$\pm$32.4} & 9.4{\small$\pm$5.0} & 70.1{\small$\pm$18.1} & 2.8{\small$\pm$1.5} \\
P10 & 3 & 23.4{\small$\pm$3.5} & 17.2{\small$\pm$1.4} & 38.3{\small$\pm$13.6} & 8.6{\small$\pm$3.3} \\
P17 & 3 & 16.1{\small$\pm$8.7} & 18.4{\small$\pm$3.2} & 46.6{\small$\pm$9.4} & 6.7{\small$\pm$0.6} \\
P19 & 3 & 22.8{\small$\pm$23.2} & 17.9{\small$\pm$8.7} & 41.0{\small$\pm$35.0} & 5.5{\small$\pm$3.1} \\
P14 & 2 & 22.6{\small$\pm$22.6} & 9.2{\small$\pm$2.1} & 71.4{\small$\pm$3.6} & 3.1{\small$\pm$0.3} \\
P16 & 2 & 34.0{\small$\pm$6.4} & 11.4{\small$\pm$2.5} & 71.3{\small$\pm$20.2} & 3.1{\small$\pm$1.1} \\
P21 & 2 & 23.5{\small$\pm$15.0} & 21.1{\small$\pm$9.0} & 57.8{\small$\pm$8.9} & 6.9{\small$\pm$2.9} \\
P36 & 2 & 87.2{\small$\pm$2.1} & 2.1{\small$\pm$0.1} & 89.4{\small$\pm$0.0} & 1.6{\small$\pm$0.2} \\
P37 & 2 & 72.3{\small$\pm$14.9} & 10.3{\small$\pm$8.6} & 50.0{\small$\pm$13.8} & 5.4{\small$\pm$2.6} \\
\midrule
\textbf{Overall} & \textbf{366} & \textbf{28.3{\small$\pm$25.0}} & \textbf{13.8{\small$\pm$7.6}} & \textbf{53.6{\small$\pm$28.7}} & \textbf{5.5{\small$\pm$3.5}} \\
\bottomrule
\end{longtable}

\clearpage
\section{Additional Qualitative Example}
\label{appendix:qualitative}

\begin{figure}[H]
    \centering
    \begin{subfigure}[b]{0.32\linewidth}
        \includegraphics[width=\linewidth]{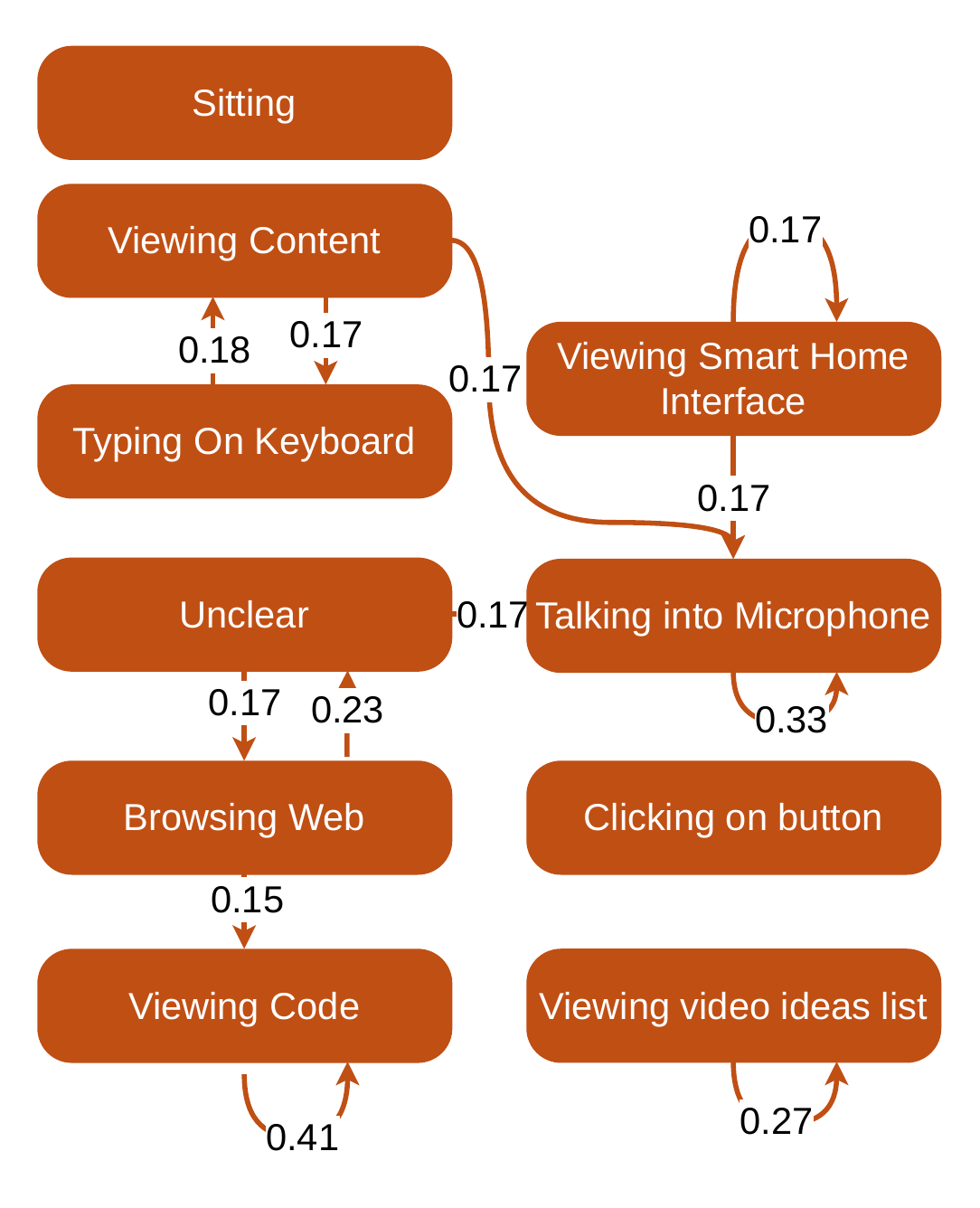}
        \caption{Pass 1 (Activity)}
    \end{subfigure}
    \hfill
    \begin{subfigure}[b]{0.33\linewidth}
        \includegraphics[width=\linewidth]{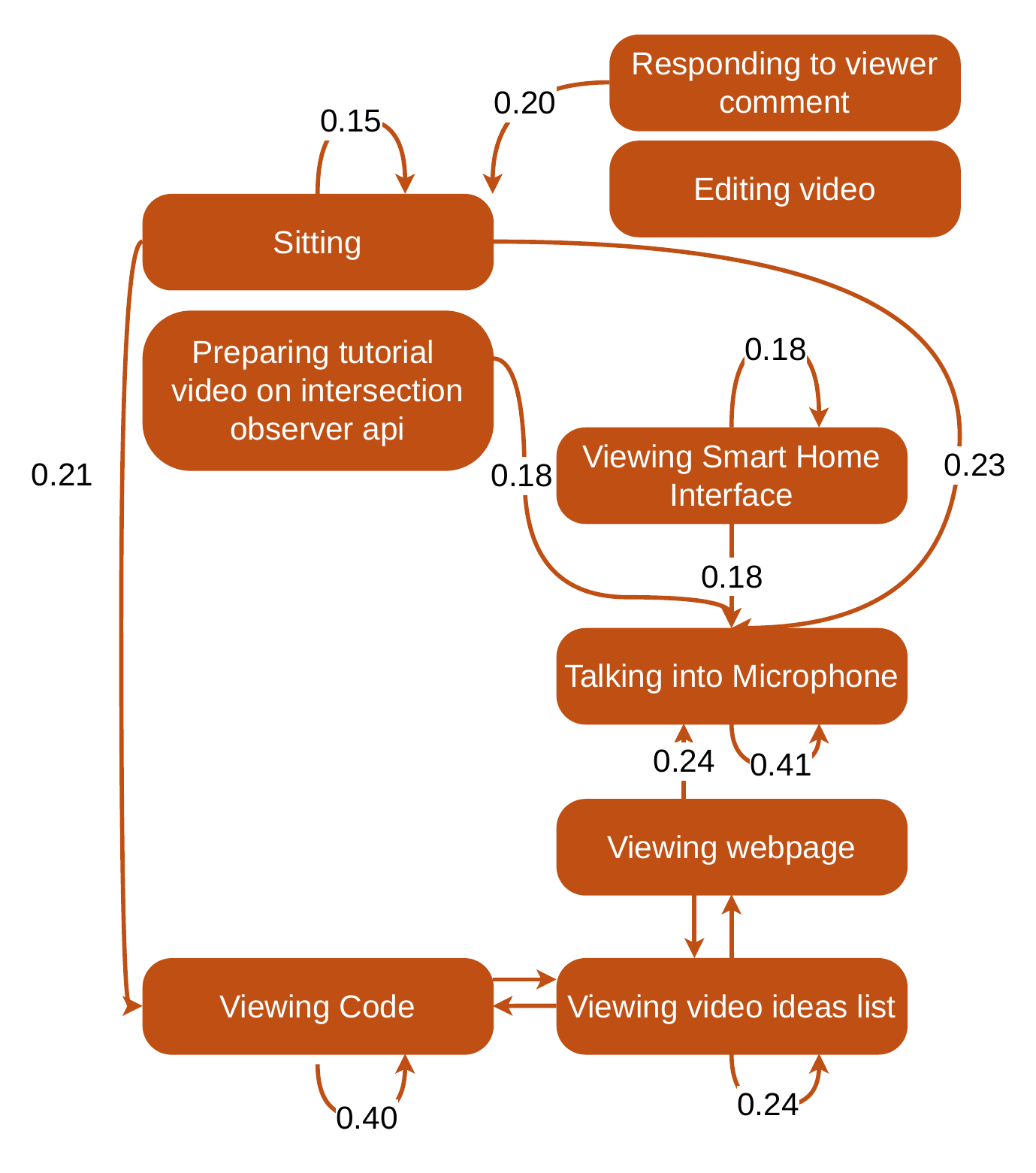}
        \caption{Pass 5 (Activity)}
    \end{subfigure}
    \hfill
    \begin{subfigure}[b]{0.33\linewidth}
        \includegraphics[width=\linewidth]{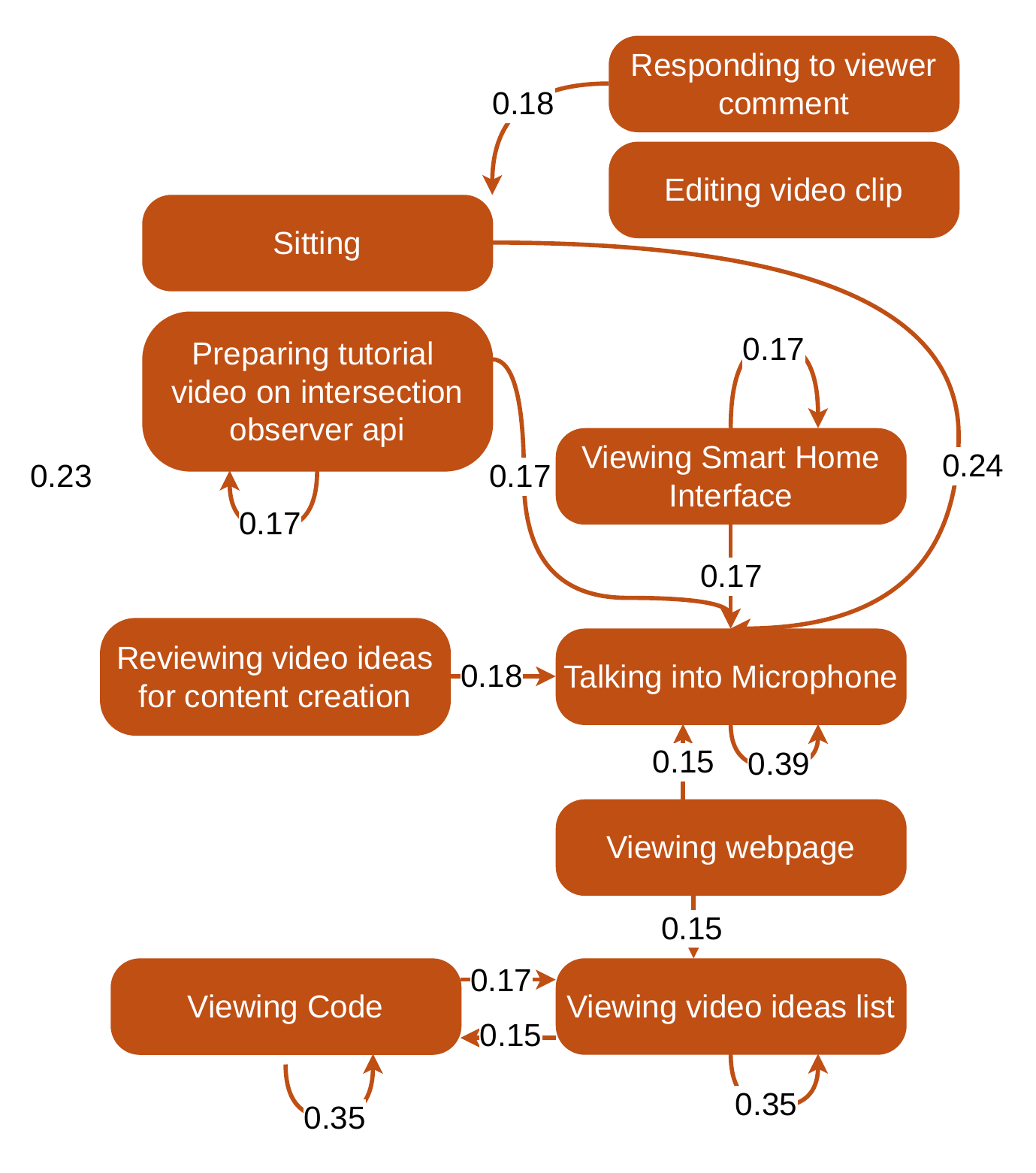}
        \caption{Pass 11 (Activity)}
    \end{subfigure}
    \\[1ex]
    \begin{subfigure}[b]{0.32\linewidth}
        \includegraphics[width=\linewidth]{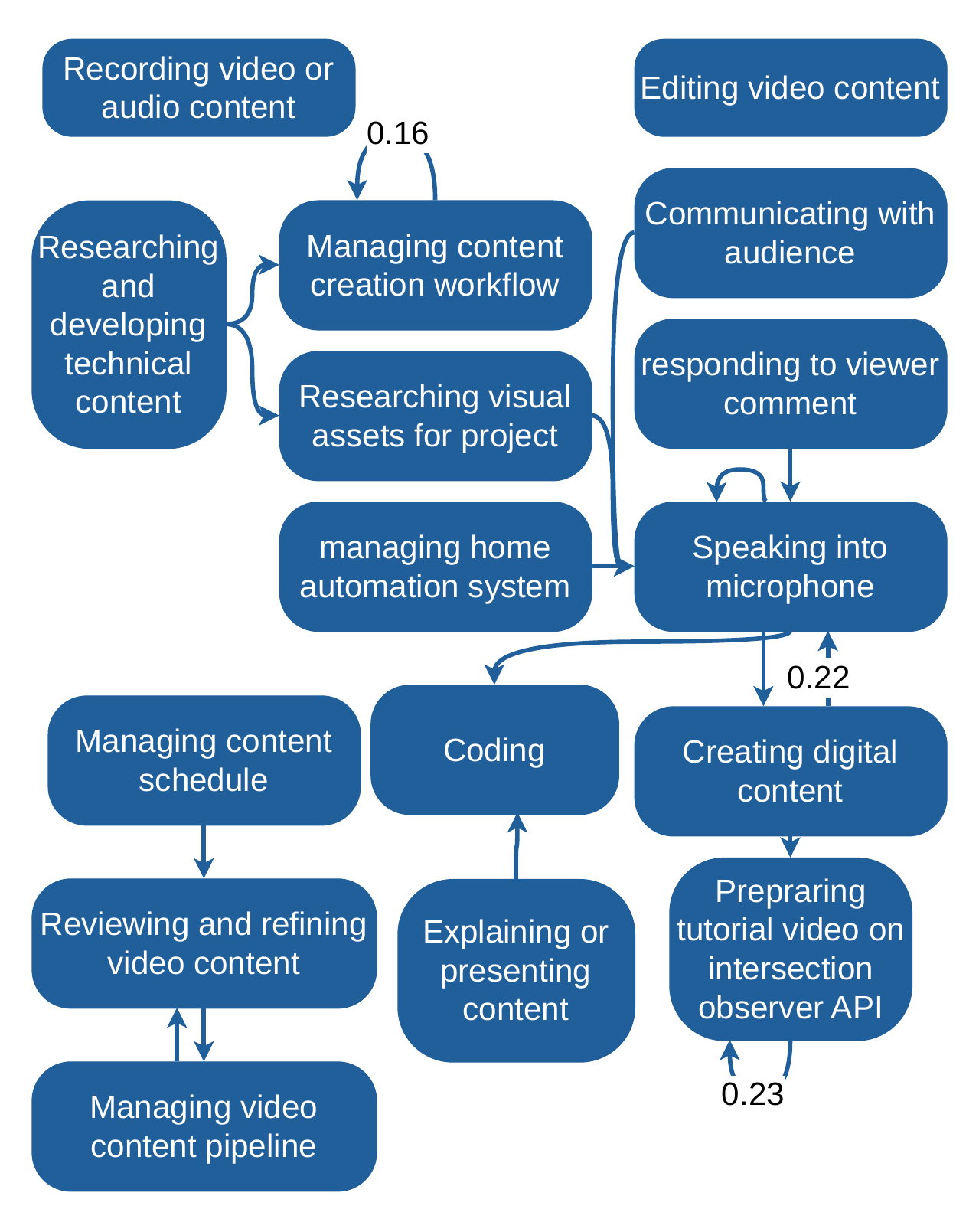}
        \caption{Pass 2 (Intent)}
    \end{subfigure}
    \hfill
    \begin{subfigure}[b]{0.33\linewidth}
        \includegraphics[width=\linewidth]{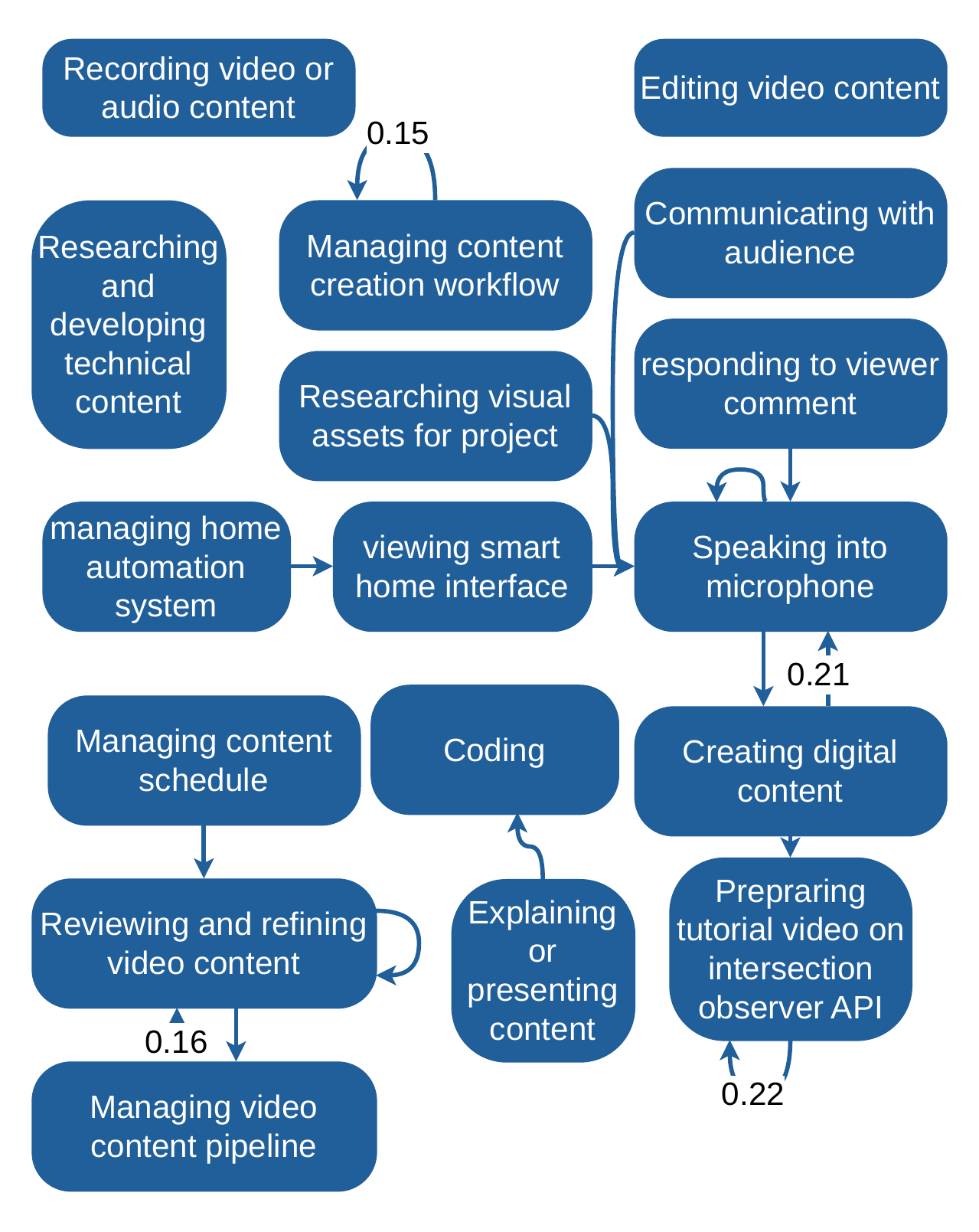}
        \caption{Pass 6 (Intent)}
    \end{subfigure}
    \hfill
    \begin{subfigure}[b]{0.33\linewidth}
        \includegraphics[width=\linewidth]{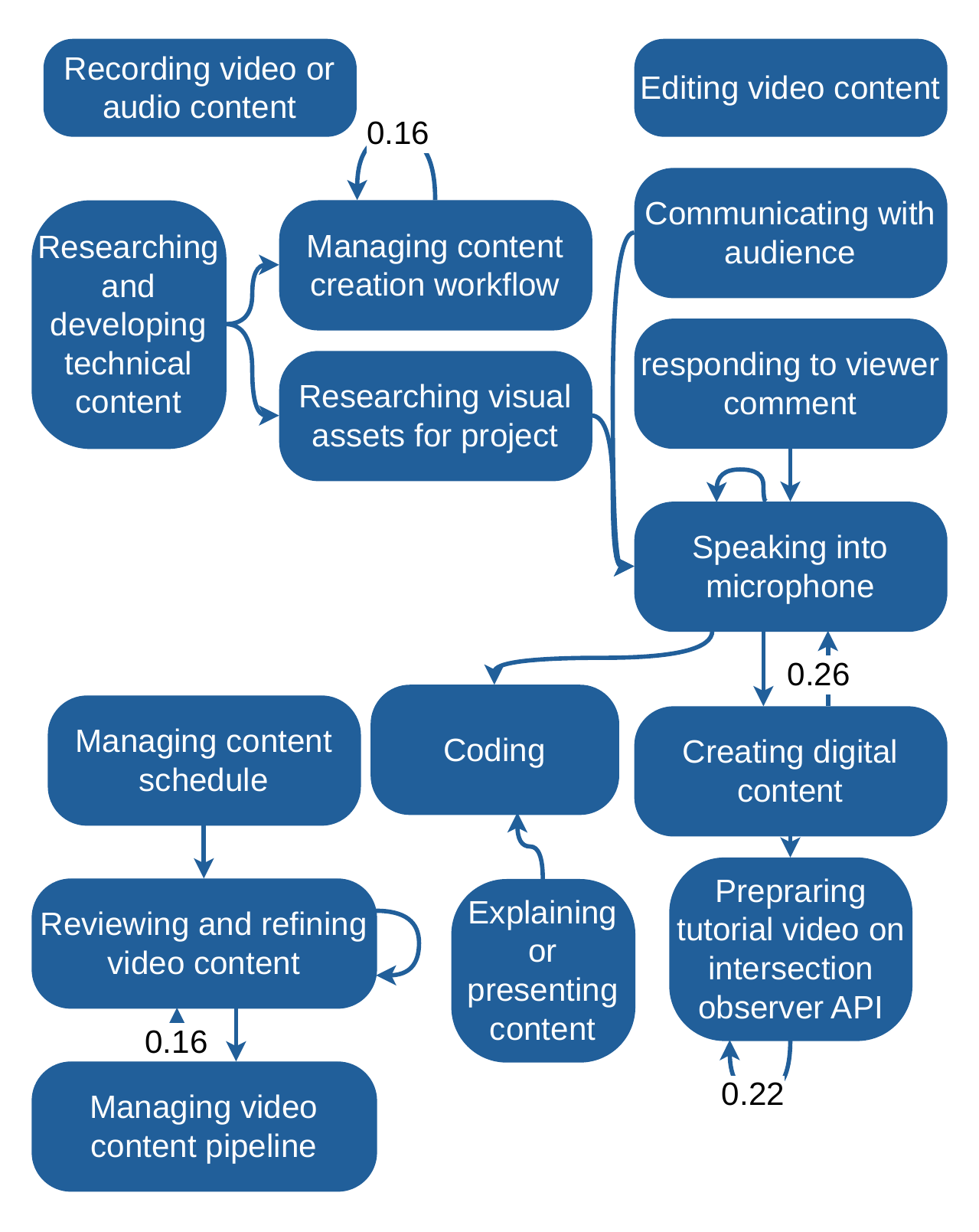}
        \caption{Pass 12 (Intent)}
    \end{subfigure}
    \caption{Content creation video (tutorialP12). \emph{Top}: activity labels refine from perceptual to task-specific. \emph{Bottom}: intent graphs reveal two workflow clusters (audience-facing vs.\ production) connected through a management hub.}
    \label{fig:passwise_markov_tutorial}
\end{figure}

\paragraph{Content creation (tutorialP12).}
Figure~\ref{fig:passwise_markov_tutorial} shows refinement on a content-creation video. Activity labels evolve from perceptual (\emph{sitting}, \emph{browsing web}) to task-specific (\emph{preparing tutorial video on intersection observer API}, \emph{responding to viewer comment}) by pass~11.

The intent graphs (d--f) reveal structure invisible in the activity graph. By pass~12, two workflow clusters emerge: an audience-facing loop (\emph{responding to viewer comment} $\to$ \emph{speaking into microphone} $\to$ \emph{creating digital content} $\to$ \emph{preparing tutorial video on intersection observer API}), and a production pipeline (\emph{managing content schedule} $\to$ \emph{reviewing and refining video content} $\to$ \emph{managing video content pipeline}). These clusters connect through \emph{managing content creation workflow}. An agent consuming this model could distinguish recording from planning phases --- a distinction the activity graph cannot surface.

\clearpage
\section{Pass-by-Pass Model Performance}
\label{appendix:pass_trajectories}

\begin{figure}[H]
    \centering
    \begin{subfigure}[b]{0.48\linewidth}
        \includegraphics[width=\linewidth]{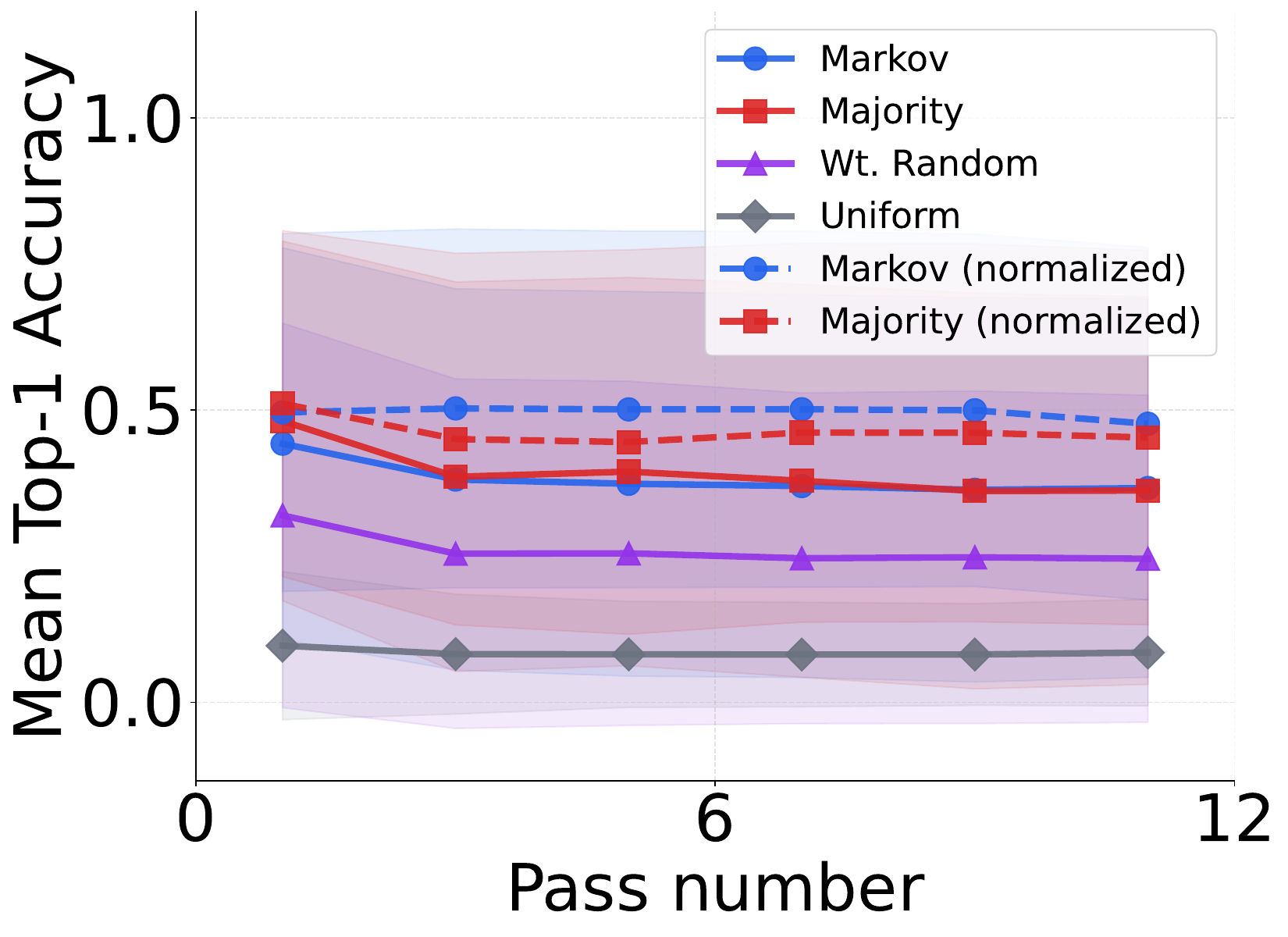}
        \caption{Activity accuracy vs baselines}
    \end{subfigure}
    \hfill
    \begin{subfigure}[b]{0.48\linewidth}
        \includegraphics[width=\linewidth]{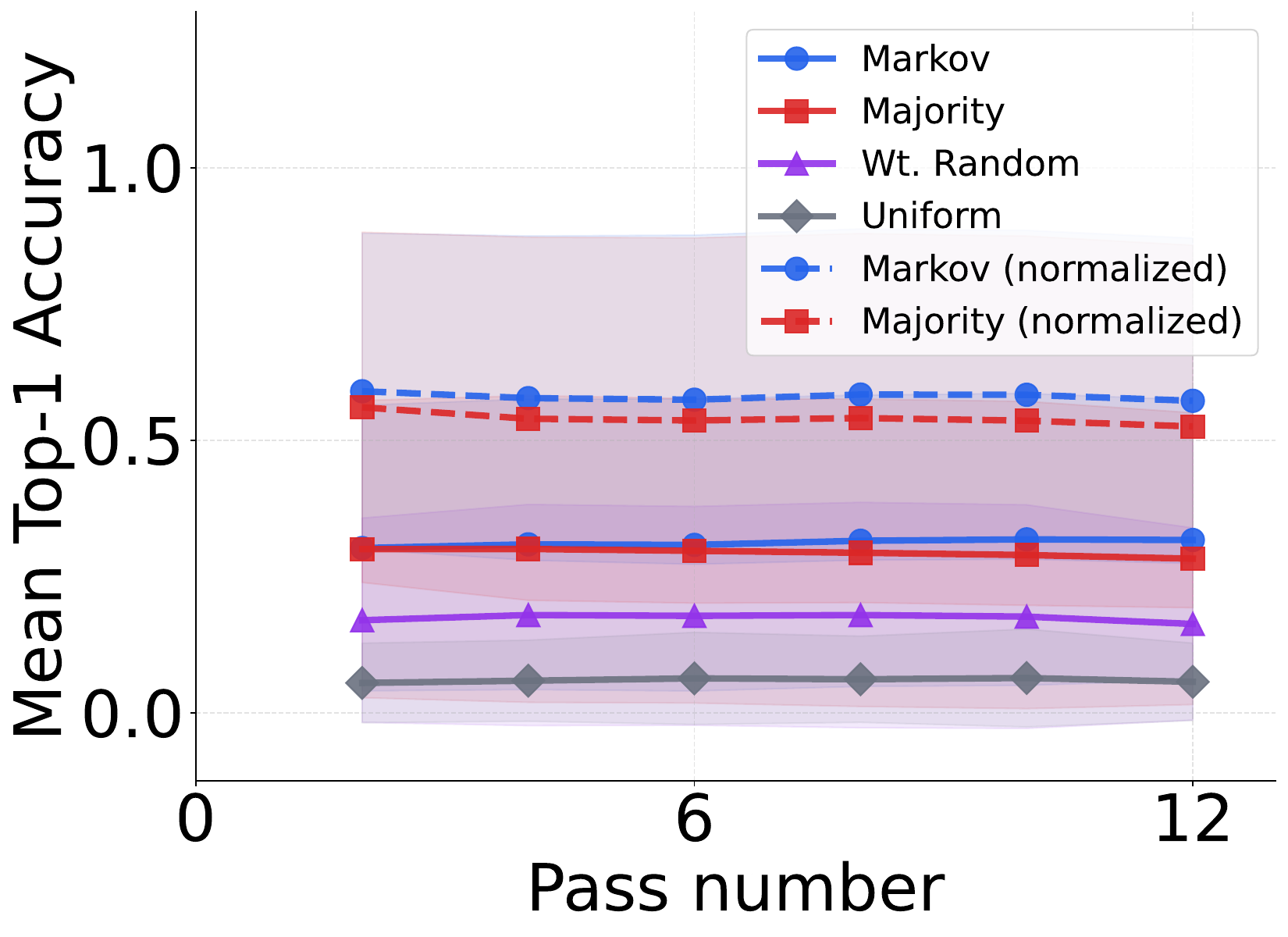}
        \caption{Intent accuracy vs baselines}
    \end{subfigure}
    \\[1ex]
    \begin{subfigure}[b]{0.48\linewidth}
        \includegraphics[width=\linewidth]{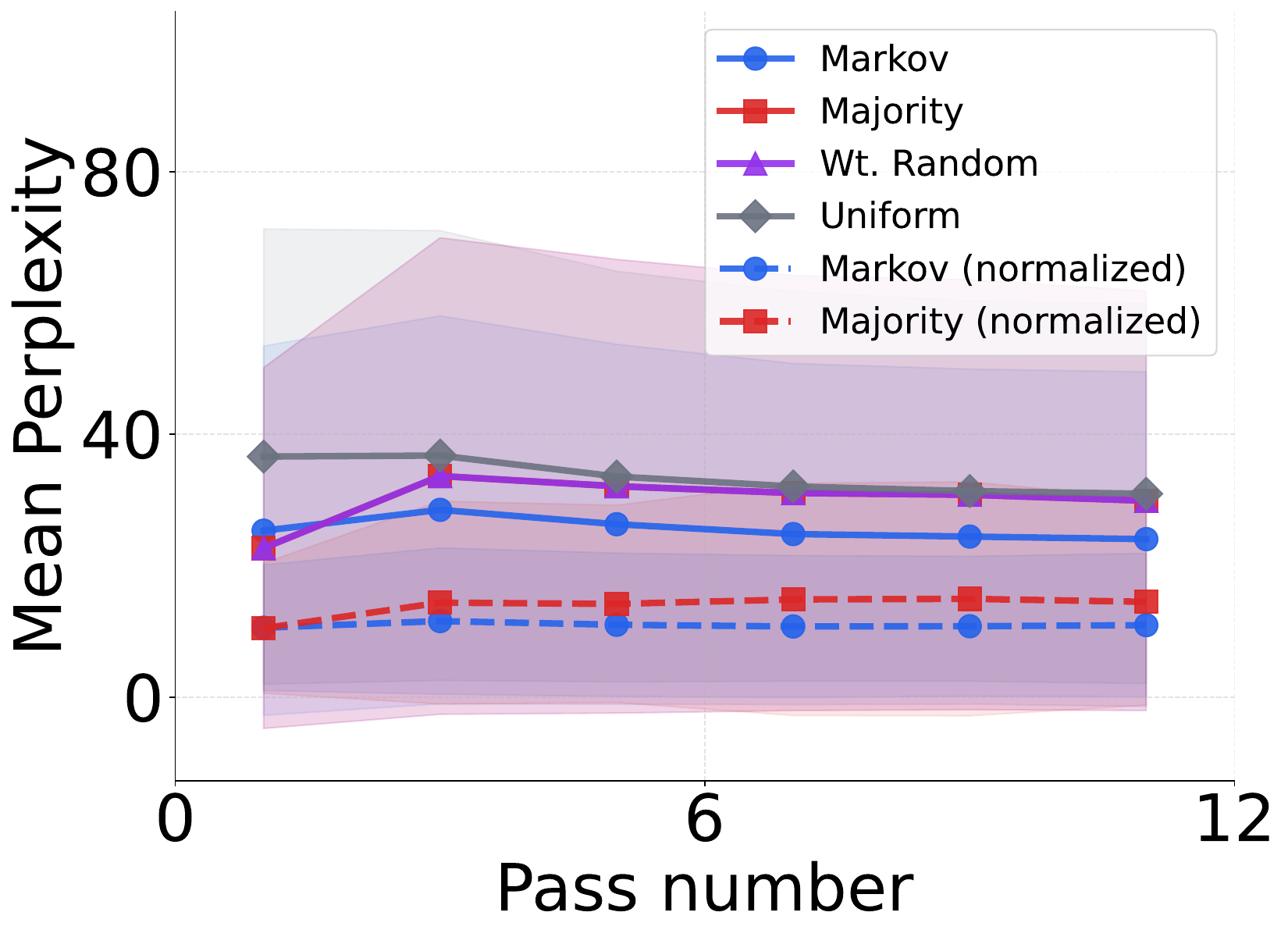}
        \caption{Activity perplexity vs baselines}
    \end{subfigure}
    \hfill
    \begin{subfigure}[b]{0.48\linewidth}
        \includegraphics[width=\linewidth]{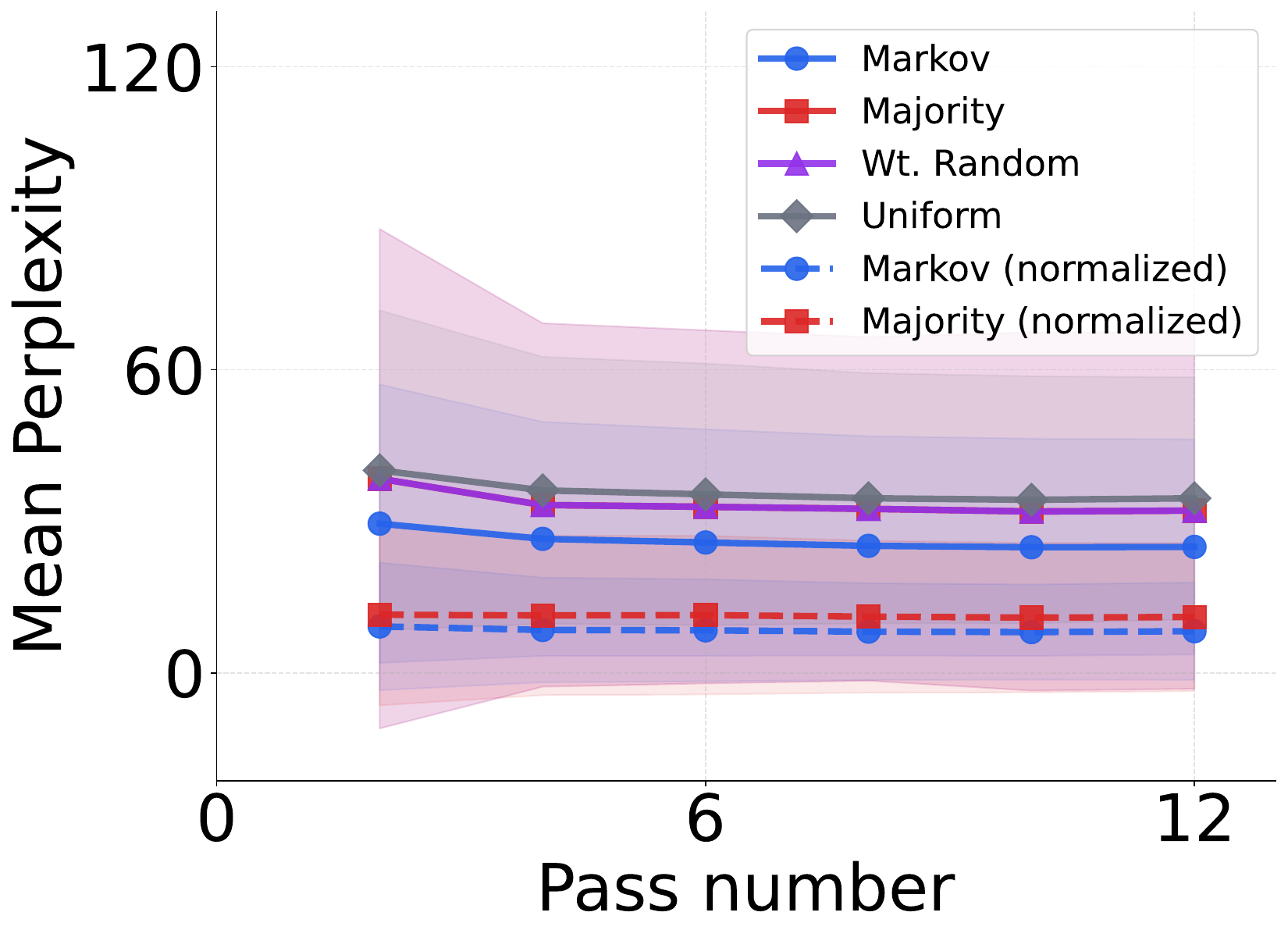}
        \caption{Intent perplexity vs baselines}
    \end{subfigure}
    \caption{Pass-by-pass top-1 accuracy and perplexity for normalized Markov vs.\ baselines across all 12 annotation passes, averaged over 61 videos with standard deviation bands.}
    \label{fig:p12_main_results}
\end{figure}

Figure~\ref{fig:p12_main_results} shows accuracy and perplexity across all 12 passes (61 videos). Activity accuracy (a): raw Markov and Majority are closely matched; normalized Markov consistently leads. Intent accuracy (b): normalized models show clearer separation, reaching ${\sim}$60\% vs.\ ${\sim}$30\% raw accuracy. Perplexity (c, d): Markov achieves the lowest at every pass; normalization roughly halves it. Wide standard deviation bands reflect high per-video variance from vocabulary size and domain differences.

\clearpage
\section{Threshold Calibration Procedure}
\label{appendix:threshold_procedure}

To calibrate the semantic merging threshold, we assemble all unique activity and intent labels across every video and pass, compute pairwise SentenceBERT cosine distances, and partition the distance range into 10 equal-width bins. We randomly sample 10 pairs per bin per type (activity, intent), yielding 100 pairs per type (200 total) stratified across the full similarity spectrum. A single annotator labels each pair as \emph{same} (semantically equivalent), \emph{different} (distinct states), or \emph{skip} (ambiguous). The optimal threshold $t^*$ is selected as the cosine distance maximizing F1 score on non-skipped pairs, treating \emph{same} as the positive class.

\clearpage
\section{Prompt Templates}
\label{appendix:prompts}

Table~\ref{tab:prompts} summarizes the four prompt templates used across annotation passes. Pass~1 receives only the frame image; all subsequent passes additionally receive the temporal context window (RLE of neighboring frames' labels) and the inter-pass summary.

\begin{table}[h]
\centering
\small
\caption{Prompt templates by pass type. Each prompt instructs the VLM to output structured JSON with a state label, confidence score (1--10), and supporting evidence.}
\label{tab:prompts}
\begin{tabular}{p{0.18\linewidth}p{0.12\linewidth}p{0.60\linewidth}}
\toprule
\textbf{Prompt} & \textbf{Passes} & \textbf{Key instruction} \\
\midrule
Activity (first) & P1 &
Identify the dominant observable action from the frame. Describe \emph{what} is happening (e.g., \texttt{typing\_on\_keyboard}), not \emph{why}. Use lowercase with underscores. \\[6pt]
\midrule
\\[-6pt]
Intent (first) & P2 &
Given prior-pass activity labels and temporal context, infer the user's underlying goal (e.g., \texttt{grocery\_shopping}). Look for patterns across sequential activities. Assign lower confidence to speculative intents. \\[6pt]
\midrule
\\[-6pt]
Refined activity & P3, P5, \ldots &
Re-analyze with enriched context from prior activity \emph{and} intent passes. Be more precise (e.g., generic \texttt{typing} + intent \texttt{coding} $\to$ \texttt{typing\_code}). Collapse synonyms. Split actions that serve different intents. Report what changed and why. \\[6pt]
\midrule
\\[-6pt]
Refined intent & P4, P6, \ldots &
Re-analyze intents with multiple passes of context. Validate or invalidate prior inferences based on subsequent observations. Discover higher-level goal patterns. Report what changed and why. \\
\bottomrule
\end{tabular}
\end{table}

Full prompt text is available in the released codebase \footnote{https://github.com/minnesotanlp/SERUM/}

\clearpage
\section{Model choice preliminary study}
\label{appendix:appendix_model_choice}

We investigated 3 new models over 4 videos chosen randomly from the expanded annotation round. We find each model still reaches schematic equilibrium at every scale tested, but generally larger models took longer to reach schematic equilibrium (Table \ref{tab:scale-trajectory}). There is no obvious pattern to the effectiveness of larger models in next-state prediction. Larger models benefit more from normalization (32B's Markov accuracy saw a 122\% increase going from 2.7 to 6.0) (Table \ref{tab:model-scale-accuracy})  primarily due to larger models being more verbose and specific about label assessments, thereby inflating vocabulary sizes (Table \ref {tab:scale-trajectory}).

\begin{table*}[h]\centering\footnotesize
\caption{Schematic equilibrium across VLM scales: vocab size by pass. 4 Qwen3-VL variants on 4 videos.}
\label{tab:scale-trajectory}
\setlength{\tabcolsep}{4pt}
\begin{tabular}{ll rrrrrr rrrrrr}
\toprule
 & & \multicolumn{6}{c}{\textbf{Intent passes}} & \multicolumn{6}{c}{\textbf{Activity passes}} \\
\cmidrule(lr){3-8}\cmidrule(lr){9-14}
Video & Model & P2 & P4 & P6 & P8 & P10 & P12 & P1 & P3 & P5 & P7 & P9 & P11 \\
\midrule
AC\_leetcode & 4B &  18 &  18 &  18 &  18 &  18 &  17 &   5 &   6 &   6 &   6 &   6 &   6 \\
 & 8B &  14 &  14 &  14 &  14 &  14 &  14 &   6 &   6 &   6 &   6 &   6 &   6 \\
 & 30B-A3B &   5 &   5 &   5 &   5 &   5 &   5 &   8 &   7 &   7 &   7 &   7 &   7 \\
 & 32B &  18 &  15 &  15 &  13 &  14 &  14 &  11 &  15 &  10 &  12 &  12 &  12 \\
\midrule
AC\_pizza & 4B &  42 &  33 &  30 &  30 &  30 &  31 &  67 &  70 &  68 &  67 &  67 &  67 \\
 & 8B &  55 &  46 &  43 &  42 &  42 &  38 &  54 &  54 &  52 &  42 &  48 &  44 \\
 & 30B-A3B &  41 &  40 &  41 &  40 &  40 &  40 &  60 &  60 &  59 &  59 &  59 &  59 \\
 & 32B &  67 &  58 &  51 &  44 &  39 &  39 &  73 &  63 &  56 &  52 &  47 &  44 \\
\midrule
AC\_ukdayinlife & 4B &  82 &  69 &  67 &  65 &  65 &  64 &  84 &  92 &  90 &  87 &  85 &  83 \\
 & 8B &  93 &  78 &  68 &  64 &  65 &  61 &  59 &  58 &  53 &  51 &  49 &  51 \\
 & 30B-A3B &  72 &  73 &  70 &  70 &  68 &  68 &  73 &  80 &  80 &  79 &  78 &  77 \\
 & 32B & 119 &  88 &  70 &  59 &  52 &  48 &  98 &  95 &  81 &  73 &  68 &  63 \\
\midrule
BC\_nycswevlog & 4B &  71 &  66 &  63 &  63 &  63 &  63 &  63 &  65 &  66 &  65 &  66 &  67 \\
 & 8B &  85 &  69 &  66 &  62 &  56 &  56 &  44 &  53 &  53 &  46 &  50 &  46 \\
 & 30B-A3B &  74 &  69 &  69 &  68 &  69 &  69 &  67 &  72 &  72 &  72 &  72 &  72 \\
 & 32B & 101 &  77 &  66 &  62 &  60 &  60 &  77 &  73 &  62 &  66 &  65 &  67 \\
\bottomrule
\end{tabular}
\end{table*}

\begin{table}[h]\centering\small
\caption{Model choice: Markov vs.\ majority accuracy (\%) and perplexity before/after normalization, by label type. $\Delta =$ Markov $-$ Majority. n = 4 videos, 4 Qwen3-VL variants.}
\label{tab:model-scale-accuracy}
\begin{tabular}{lrrrrrrrr}
\toprule
 & \multicolumn{4}{c}{Raw} & \multicolumn{4}{c}{Normalized} \\
Model & Markov & Maj. & $\Delta$ & PPL & Markov & Maj. & $\Delta$ & PPL \\
\midrule
\multicolumn{9}{l}{\textit{Activity (P11)}} \\
4B & 25.7 & 27.4 & -1.8 & 47.6 & 25.8 & 30.2 & -4.4 & 28.3 \\
8B & 32.1 & 30.3 & +1.8 & 30.4 & 33.5 & 29.5 & +4.0 & 20.5 \\
30B-A3B & 17.4 & 12.3 & +5.2 & 49.3 & 21.0 & 12.8 & +8.2 & 28.8 \\
32B & 2.7 & 4.5 & -1.8 & 44.0 & 6.0 & 6.8 & -0.7 & 28.2 \\
\midrule
\multicolumn{9}{l}{\textit{Intent (P12)}} \\
4B & 10.0 & 6.6 & +3.4 & 39.8 & 17.4 & 8.8 & +8.6 & 24.6 \\
8B & 23.7 & 21.4 & +2.2 & 36.9 & 26.5 & 26.9 & -0.4 & 22.4 \\
30B-A3B & 28.8 & 33.0 & -4.3 & 37.7 & 38.8 & 41.0 & -2.2 & 18.7 \\
32B & 5.7 & 8.7 & -3.0 & 37.8 & 10.1 & 8.2 & +1.9 & 23.5 \\
\bottomrule
\end{tabular}
\end{table}
\clearpage
\section{Transferability Study}
\label{appendix:transferability_study}

As a preliminary check on cross-video transferability, we ran 3 leave-one-out markov evaluations on two same domain video triples: car repair, and coffee shop operation each (6 total). For car repair videos, Markov improved over the Majority baseline on the held-out video, demonstrating the transferability of learned models to new videos. On the other hand, in the coffee shop triple, one state (pouring\_milk\_into\_cup) occurs very frequently (27\%-39\% frames in each video); in this case, Majority demonstrates better transferability.

\begin{table}[h]\centering\small
\caption{Cross-video Markov transferability (LOOCV): train on 2 videos' concatenated activity sequences, test on the held-out video. Normalized vocabulary on the union of each triple at $t^*=0.43$.}
\label{tab:transferability-pilot}
\begin{tabular}{llrrrr}
\toprule
Triple & Held-out test & Novel \% & Markov & Majority & Markov $-$ Majority \\
\midrule
\multicolumn{6}{l}{\textit{Car repair}} \\
 & \texttt{carrepair} & 53.8\% & 15.3\% & 1.8\% & +13.5 \\
 & \texttt{carrepair2} & 53.1\% & 22.3\% & 8.0\% & +14.3 \\
 & \texttt{carrepair3} & 38.6\% & 14.1\% & 0.0\% & +14.1 \\
 & \textit{Mean} & --- & --- & --- & \textit{+14.0} \\
\midrule
\multicolumn{6}{l}{\textit{Coffee shop}} \\
 & \texttt{dunkin} & 24.8\% & 21.1\% & 27.3\% & -6.2 \\
 & \texttt{BC\_dunkinhelp} & 23.0\% & 25.9\% & 38.7\% & -12.7 \\
 & \texttt{CC\_barista} & 38.7\% & 22.2\% & 28.5\% & -6.3 \\
 & \textit{Mean} & --- & --- & --- & \textit{-8.4} \\
\bottomrule
\end{tabular}
\end{table}

We were also interested if there was significant pair-wise video transferability and conducted a brief pairwise transferability experiment, and found that yes if two videos are similar enough they are transferable; label ontology is relatively consistent for similar videos.

\begin{table}[t]\centering\small
\caption{Cross-video Markov transferability (pairwise): each video used as a single training set against another single video as test. More conservative than LOOCV (less train data, higher novel \%).}
\label{tab:transferability-pairwise}
\begin{tabular}{llrrrr}
\toprule
Train & Test & Novel \% & Markov & Majority & Markov $-$ Majority \\
\midrule
\multicolumn{6}{l}{\textit{Car repair}} \\
\texttt{carrepair} & $\to$ \texttt{carrepair2} & 53.1\% & 19.6\% & 1.8\% & +17.9 \\
\texttt{carrepair} & $\to$ \texttt{carrepair3} & 38.6\% & 7.6\% & 5.3\% & +2.4 \\
\texttt{carrepair2} & $\to$ \texttt{carrepair} & 69.5\% & 10.4\% & 0.0\% & +10.4 \\
\texttt{carrepair2} & $\to$ \texttt{carrepair3} & 66.7\% & 8.2\% & 0.0\% & +8.2 \\
\texttt{carrepair3} & $\to$ \texttt{carrepair} & 60.1\% & 10.4\% & 1.8\% & +8.6 \\
\texttt{carrepair3} & $\to$ \texttt{carrepair2} & 69.0\% & 13.4\% & 8.0\% & +5.4 \\
\textit{Mean} & & --- & --- & --- & \textit{+8.8} \\
\midrule
\multicolumn{6}{l}{\textit{Coffee shop}} \\
\texttt{dunkin} & $\to$ \texttt{BC\_dunkinhelp} & 31.5\% & 27.4\% & 38.7\% & -11.3 \\
\texttt{dunkin} & $\to$ \texttt{CC\_barista} & 45.0\% & 21.3\% & 28.5\% & -7.2 \\
\texttt{BC\_dunkinhelp} & $\to$ \texttt{dunkin} & 35.2\% & 21.1\% & 27.3\% & -6.2 \\
\texttt{BC\_dunkinhelp} & $\to$ \texttt{CC\_barista} & 47.8\% & 21.0\% & 28.5\% & -7.5 \\
\texttt{CC\_barista} & $\to$ \texttt{dunkin} & 36.7\% & 19.1\% & 27.3\% & -8.1 \\
\texttt{CC\_barista} & $\to$ \texttt{BC\_dunkinhelp} & 32.9\% & 25.5\% & 38.7\% & -13.2 \\
\textit{Mean} & & --- & --- & --- & \textit{-8.9} \\
\bottomrule
\end{tabular}
\end{table}

\clearpage

\section{LLM Disclosure}
\label{appendix:appendix_llm_disclosure}

In accordance with the COLM 2026 policy on LLM usage, we disclose the following. 
LLM-assisted coding tools were used during software development and infrastructure 
management. An LLM was also used to proofread drafts and assist with an initial 
literature survey; all references were verified by the authors. 

LLMs were not used to generate experimental results, figures, datasets, or 
quantitative analysis. The research ideas, experimental design, implementation, 
analysis, and paper content are the work of the authors.

\end{document}